\documentclass{article}

\usepackage{microtype}
\usepackage{graphicx}
\usepackage{subfigure}
\usepackage{booktabs} % for professional tables
\usepackage{framed,color,verbatim}
\definecolor{shadecolor}{rgb}{.9, .9, .9}
\usepackage{fancyvrb}
\usepackage{xcolor}
\usepackage{empheq}
\usepackage{hyperref}

\usepackage{framed,color,verbatim}
\definecolor{shadecolor}{rgb}{.9, .9, .9}
\definecolor{shadecolor}{rgb}{.1, .9, .1}

\usepackage{tikz}
\usetikzlibrary{decorations.pathreplacing,calc}
\usepackage{enumitem}

\usepackage[most]{tcolorbox}
\usepackage{xcolor}         % colors

\tcbset{%
    theo/.style={%
        enhanced,
        sharp corners,
        toprule=0pt, rightrule=0pt, bottomrule=0pt, leftrule=1mm,
        colback=#1!10, colframe=#1!80!black, coltitle=#1!0!black, 
        left=2mm, right=2mm, top=1mm, bottom=2mm, % Adjust the margins
        detach title,
        overlay unbroken and first ={
            \node[rotate=90, minimum width=1cm, anchor=south, font=\bfseries] 
               at (frame.west) {\tcbtitle};
        },
    }
}
\newtcbtheorem[auto counter]{curlybracketwhite}{Attend}{theo=white}{th}
\newtcbtheorem[auto counter]{curlybracketgrey}{Don't attend}{theo=green}{th}
\newtcbtheorem[auto counter]{curlybracketgreen}{Attend}{theo=green}{th}

\tcbset{%
    no_side_note/.style={%
        enhanced,
        sharp corners,
        toprule=-7.5pt, rightrule=0pt, bottomrule=0pt, leftrule=0pt,
        colback=#1!5, colframe=#1!80!black, coltitle=#1!0!black, 
        left=0mm, right=2mm, top=1mm, bottom=1mm, % Adjust the margins
    }
}
\newtcbtheorem[auto counter]{plainbox}{}{no_side_note=green}{th}

\usepackage{multirow}

   {\snugshade\verbatim}%
   {\endverbatim\endsnugshade}

\newcommand{\cut}[1]{}
\usepackage[accepted]{icml2021}

\icmltitlerunning{Programming in GPT}

\begin{document}

\twocolumn[
\icmltitle{GPT is becoming a Turing machine: \\
           Here are some ways to program it}




\begin{icmlauthorlist}
\icmlauthor{Ana Jojic}{fred}
\icmlauthor{Zhen Wang}{UCSD}
\icmlauthor{Nebojsa Jojic}{msr}
\end{icmlauthorlist}

\icmlaffiliation{fred}{Fred Hutchinson Cancer Research Center, Seattle, WA, USA}
\icmlaffiliation{msr}{Microsoft Research, Redmond, WA, USA}
\icmlaffiliation{UCSD}{University of California San Diego, La Jolla, CA, USA, and Mohamed bin Zayed University of Artificial Intelligence, Masdar City, Abu Dhabi}

\icmlcorrespondingauthor{Nebojsa Jojic}{jojic@microsoft.com}

\icmlkeywords{Machine Learning, ICML}

\vskip 0.3in
]



\printAffiliationsAndNotice{}  % leave blank if no need to mention equal contribution

\begin{abstract}
We demonstrate that, through appropriate prompting, GPT-3 family of models can be triggered to perform iterative behaviours necessary to execute (rather than just write or recall) programs that involve loops, including several popular algorithms found in computer science curricula or  software developer interviews.  We trigger execution and description of {\bf iterations} by {\bf regimenting self-attention} (IRSA)  in one (or a combination) of three ways: 1) Using strong repetitive structure in an example of an execution path of a target program for one particular input, 2) Prompting with fragments of execution paths, and 3) Explicitly forbidding (skipping) self-attention to parts of the generated text. On a dynamic program execution, IRSA leads to larger accuracy gains than replacing the model with the much more powerful GPT-4. IRSA has promising applications in education, as the prompts and responses resemble student assignments in data structures and algorithms classes. Our findings hold implications for evaluating LLMs, which typically target the in-context learning: We show that prompts that may not even cover one full task example can trigger algorithmic behaviour, allowing solving problems previously thought of as hard for LLMs, such as logical puzzles. Consequently, prompt design plays an even more critical role in LLM performance than previously recognized.
\end{abstract}

\section{Introduction}

Large language  models (LLMs) \cite{GPT3,rae2021scaling,chowdhery2022palm, openai2023gpt4} are trained on large amounts of text data, which typically include descriptions of procedures and even computer programs~\cite{copilot}. They have demonstrated a surprisingly high competency in retrieving knowledge from the training data and generalizing it to new, slightly different situations. The models are typically evaluated on ``in-context learning'' tasks, i.e., zero- and few-shot prompting, with results implying that these models compress iterative reasoning into a savant-like ability to directly reach correct conclusions without a disciplined step-by-step process~\cite{wei2022chain, kojima2022large}. It is difficult to understand if these abilities are simply due to a high similarity with the training data, or if they are evidence of the ever-increasing generalization. 

In practice, however, even in simple scenarios where the justification of answers to a given question requires a couple of reasoning steps, providing those steps in the prompt for a few examples improves the accuracy of LLMs. Early such approaches include \cite{shwartz-etal-2020-unsupervised,zelikman2022star,nye2021show}, while more general Chain-of-Thought (CoT) prompting methods include \citep{wei2022chain, wang2022self, zhou2022least, creswell2022selection, wang2022rationale,liu-etal-2022-multi, kojima2022large, li2022advance}. This implies that despite the massive number of parameters and the self-attention to all previous tokens, current LLMs are unlikely to solve problems that require many (or iterated) reasoning steps in a direct, savant-like manner. In designing new benchmarks, the NLP community has been targeting more complex tasks where humans would not only need detailed reasoning to justify their answer, but need it to reach the conclusions in the first place. Several tasks, such as logical deduction and logical grid puzzles in BIG-bench Lite \citep{srivastava2022beyond}, require constraint satisfaction propagation to solve, and in-context learning of these problems is typically poor. 
LLMs excite us with apparent emergence of such savant abilities elsewhere, as evidenced by GitHub Copilot usage statistics~\cite{copilot_study}, where nearly 50\% of code is auto-generated by Codex~\cite{copilot}. But Copilot is a human-in-the-loop application, and standalone systems like AlphaCode~\cite{alphacode} and Codex fall short compared to expert programmers, mostly because they guess the answer but have no ability to execute, track state, and debug programs (apart from anecdotal evidence, e.g. Fig. 3.7 in \cite{bubeck2023sparks}; also see Section \ref{sec:discussion}).

LLMs simply generate tokens in order, each based on many (ideally all) previous tokens in the sequence, whether these tokens were part of the prompt or were just generated by the LLM itself. Thus, the self-attention could allow an LLM to use all previously generated tokens as the scratchpad for tracking reasoning steps, states, etc\footnote{This is likely to be one of the reasons for the increased performance of CoT prompting.}. Such a use of generated tokens would resemble a classical Turing Machine with its memory tape \cite{turing1936a}. In principle, a non-trivial recurrent transformer model with infinite attention could be Turing-complete and capable of executing arbitrary routines, as long as the attention mechanism can be controlled stringently enough. But, even in relatively simple settings, LLMs appear to resist strict controls, e.g., slight changes in prompts can yield dramatically different responses~\cite{liu2021makes, malkin2022coherence, shi2023large}, because many recurrent patterns in the training data are encoded into a single model, and learned patterns overlap and vary in the context size. Thus it is easy to mislead with a prompt that has accidental alphabetical or numerical ordering, or happens to deal only with one type of object, etc~\citep{zhao2021calibrate, lu-etal-2022-fantastically, min2022rethinking}. 

In this paper, we study much stricter attention controls that instruct LLMs to unroll the reasoning steps over a long procedure with the initially undetermined length, decide when the solution was found, and output the result. A simple example is given in Prompt \ref{pr:BS}, where a prompt describing the state evolution of the Bubble Sort algorithm on one example turns GPT-3 into a Bubble Sorter that outputs not only the sorted sequence but other details of the state, including the number of element swaps performed (response in Prompt \ref{pr:BSresp}). While the latest LLMs can sort lists and even recall the Bubble Sort algorithm in multiple programming languages, inferring the number of swaps, as a measure of sequence disorder, is one type of difficult reasoning task that is hard to solve in a savant manner. This example alone immediately points to important consequences discussed in Section \ref{sec:discussion}. First, LLMs can (or soon will be able to) execute arbitrary code and thus can have applications beyond Copilot in software engineering and education~\cite{gao2022pal, parisi2022talm, schick2023toolformer, mialon2023augmented}. More pressingly, our findings point out the potentially problematic issue of in-context learning evaluation on LLMs: if prompts can combine typical natural language instructions with algorithmic iterative reasoning, then abilities of LLMs in zero- and few-shot learning with CoT reasoning are vastly underestimated, and their comparisons are deceptive. The sensitivity of the performance to prompt design may even be amplified by the iterative reasoning triggered by the prompt, which will then beg the question: If one LLM beats the other on a task, is it simply because we have not found the right prompt for the second model? For example, Prompt \ref{pr:LD} increases the performance of GPT-3 family on logical deduction puzzles (Section \ref{sec:LD}) from $32\%$ to $76\%$ (Table \ref{tab:main}, Section \ref{sec:LD}). 

While examples in Prompts \ref{pr:BS} and \ref{pr:LD} may already be sufficient to make these points, we show that these are not one-off examples by presenting more detailed results on a wider range of algorithms taught in computer science curricula and used to test software engineers in coding interviews, including string manipulations, dynamic programming, and stack operations. In Section \ref{sec:irsa}, we discuss how we were able to trigger the execution (rather than just recollection) of algorithms consistently, across varied inputs, through deliberate control of self-attention, a technique we refer to as Iteration by Regimenting Self-Attention (IRSA). The basic way to achieve this, demonstrated in Prompts \ref{pr:BS}, \ref{pr:LD}, \ref{pr:BS2}, \ref{pr:lss}, and \ref{pr:para}, is through highly structured prompting with an example of an execution path for one example. We also investigate prompting that combines multiple fragments of execution paths in Section \ref{sec:fragments}, and a strategy of skipping parts of generated text when performing self-attention in Section \ref{sec:skip}. In Section \ref{sec:interpret}, we show how one may use these ideas to further design interpreter/compiler prompts that can translate an algorithm in a high-level programming language into an IRSA prompt that GPT-3 can execute. Section \ref{sec:results} contains numerical results on several datasets from BIG-bench and tasks of our own-making, and finally, the discussion in Section \ref{sec:discussion} covers some lessons learned on how hard it is to ``program'' in GPT as ``machine code,'' as well as possible practical consequences of our findings. The discussion also includes an experiment with GPT-4 \cite{openai2023gpt4} on a well known dynamic programming task showing that even the latest (at the moment) member in the family cannot consistently execute code without more careful prompting in IRSA style.

\section{Iteration by Regimenting Self Attention (IRSA)}
\label{sec:irsa}

\begin{prompt*}
    \caption{Bubble Sort: The prompt describes iterative state evolution, including counting swaps, and making the deterimination when to stop. \href{https://platform.openai.com/playground/p/TBXnrq1eEreegSknw8wpXh9e?model=code-davinci-002}{Playground link (use with 0 temperature)}}
    \label{pr:BS}
    \begin{small}
        \begin{verbatim}
Problem: 2, 3, 1, 5
EXECUTION
    Prep
    Length of the list: 4
    Number of consecutive pairs: 3
    a=[2 3 1 5]
    set n_swaps=0
    EndPrep
    Iteration:
       set swap_flag=false. The state is:
       State: a=[2 3 1 5], n_swaps=0, swap_flag=false EndState
       Pair a[1,2] = [2 3] Check if 2<3. Is it true? Yes.
                           Because of that, we leave state as is
       State: a=[2 3 1 5], n_swaps=0, swap_flag=false
       Pair a[2,3]= [3  1] Check if 3<1.  Is it true? No.
                           Thus, we swap_flag=true, increase n_swaps by one, 
                           and in the latest a=[2 3 1 5] swap 3 and 1 to get into state:
       State: a=[2 1 3 5], n_swaps=1, swap_flag=true EndState
       Pair a[3,4]= [3 5]  Check if 3<5. Is it true? Yes.
                           Because of that, we leave state as is
       State: a=[2 1 3 5], n_swaps=1, swap_flag=true EndState
       swap_flag is true, so do another iteration
    Iteration:
       set swap_flag=false. The state is:
       State: a=[2 1 3 5], n_swaps=1, swap_flag=false EndState
       Pair a[1,2] = [2 1] Check if 2<1. Is it true? No.
                           Thus, we set swap_flag=true, increase n_swaps by one,
                           and in the latest a=[2, 1, 3, 5] swap 2 and 1 to get into state:
       State: a=[1 2 3 5], n_swaps=2, swap_flag=true EndState
       Pair a[2,3] = [2 3] Check if 2<3. Is it true? Yes.
                           Because of that, we leave state as is
       State: a=[1 2 3 5], n_swaps=2, swap_flag=true EndState
       Pair a[3,4] = [3 5] Check if 3<5. Is it true? Yes.
                           Because of that, we leave state as is
       State: a=[1 2 3 5], n_swaps=2, swap_flag=true EndState
       swap_flag is true, so do another iteration
    Iteration:
       set swap_flag=false. The state is:
       State: a=[1 2 3 5], n_swaps=2, swap_flag=false EndState
       Pair a[1,2] = [1 2] Check if 1<2. Is it true? Yes.
                           Because of that, we leave state as is
       State: a=[1 2 3 5], n_swaps=2, swap_flag=false EndState
       Pair a[2,3] = [2 3] Check if 2<3. Is it true? Yes.
                           Because of that, we leave state as is
       State: a=[1 2 3 5], n_swaps=2, swap_flag=false EndState
       Pair a[3,4] = [3 5] Check if 3<5. Is it true? Yes.
                           Because of that, we leave state as is
       State: a=[1 2 3 5], n_swaps=2, swap_flag=false EndState
       swap_flag is false, so stop the iteration
Final List: 1, 2, 3, 5
Number of swaps: 2
END OF EXECUTION

Problem: 0, 3, 8, 5, 6
EXECUTION
\end{verbatim}
\end{small}
\end{prompt*}

\begin{prompt*}
\caption{Response (shaded green) to Prompt \ref{pr:BS}, whose ending part is shown.}
\label{pr:BSresp}
    \begin{small}
\begin{verbatim}

...
       State: a=[1 2 3 5], n_swaps=2, swap_flag=false EndState
       swap_flag is false, so stop the iteration
Final List: 1, 2, 3, 5
Number of swaps: 2
END OF EXECUTION

Problem: 0, 3, 8, 5, 6
EXECUTION
\end{verbatim}
%\definecolor{shadecolor}{rgb}{.835, .957, .824}
% \definecolor{shadecolor}{rgb}{0.9569, 1.0000, 0.9490}
\definecolor{shadecolor}{rgb}{0.902, 1, 0.902}
\begin{shaded}
% \begin{plainbox*}{}{}
\begin{verbatim}
    Prep
    Length of the list: 5
    Number of consecutive pairs: 4
    a=[0 3 8 5 6]
    set n_swaps=0
    EndPrep
    Iteration:
       set swap_flag=false. The state is:
       State: a=[0 3 8 5 6], n_swaps=0, swap_flag=false EndState
       Pair a[1,2] = [0 3] Check if 0<3. Is it true? Yes.
                           Because of that, we leave state as is
       State: a=[0 3 8 5 6], n_swaps=0, swap_flag=false EndState
       Pair a[2,3] = [3 8] Check if 3<8. Is it true? Yes.
                           Because of that, we leave state as is
       State: a=[0 3 8 5 6], n_swaps=0, swap_flag=false EndState
       Pair a[3,4] = [8 5] Check if 8<5. Is it true? No.
                           Thus, we set swap_flag=true, increase n_swaps by one,
                           and in the latest a=[0 3 8 5 6] swap 8 and 5 to get into state:
       State: a=[0 3 5 8 6], n_swaps=1, swap_flag=true EndState
       Pair a[4,5] = [8 6] Check if 8<6. Is it true? No.
                           Thus, we set swap_flag=true, increase n_swaps by one,
                           and in the latest a=[0 3 5 8 6] swap 8 and 6 to get into state:
       State: a=[0 3 5 6 8], n_swaps=2, swap_flag=true EndState
       swap_flag is true, so do another iteration
    Iteration:
       set swap_flag=false. The state is:
       State: a=[0 3 5 6 8], n_swaps=2, swap_flag=false EndState
       Pair a[1,2] = [0 3] Check if 0<3. Is it true? Yes.
                           Because of that, we leave state as is
       State: a=[0 3 5 6 8], n_swaps=2, swap_flag=false EndState
       Pair a[2,3] = [3 5] Check if 3<5. Is it true? Yes.
                           Because of that, we leave state as is
       State: a=[0 3 5 6 8], n_swaps=2, swap_flag=false EndState
       Pair a[3,4] = [5 6] Check if 5<6. Is it true? Yes.
                           Because of that, we leave state as is
       State: a=[0 3 5 6 8], n_swaps=2, swap_flag=false EndState
       Pair a[4,5] = [6 8] Check if 6<8. Is it true? Yes.
                           Because of that, we leave state as is
       State: a=[0 3 5 6 8], n_swaps=2, swap_flag=false EndState
       swap_flag is false, so stop the iteration
Final List: 0, 3, 5, 6, 8
Number of swaps: 2
END OF EXECUTION
\end{verbatim}
% \end{plainbox*}
\end{shaded}
\end{small}
\end{prompt*}

Prompts \ref{pr:BS}, \ref{pr:LD}, \ref{pr:BS2}, \ref{pr:lss}, and \ref{pr:para} show the basic IRSA. In each of these examples, a single prompt is provided for a task, which, when combined with a new instance of the task, trigger the execution of an iterative algorithm. The algorithms are single loop (Prompts \ref{pr:lss} and \ref{pr:para}) or double loop (Prompts \ref{pr:BS}, \ref{pr:BS2}, and \ref{pr:LD}), and may have a known or unknown number of iterations until termination.

Crucially, the prompts show all state changes and \emph{explain each change before it occurs}.
% \emph{but before each change, explain why it is being made}. 
Although the explanation is colloquial, the structure of it is both rigid and repetitive, strictly regimenting the attention to the rules (corresponding to program instructions) and state changes. In all these examples, this strategy hardens the attention sufficiently to facilitate disciplined procedural reasoning, while leaving non-regimented content open to interpretation. For example, Prompt \ref{pr:BS} shows how a sequence of 4 integers can be sorted in some detail, but the same prompt can also be used to sort characters alphabetically or animals by size, and the procedure typically works for both shorter and longer lists. More on this in Section \ref{sec:LD}.

These prompts could be thought of as an instance of Chain-of-Thought prompting \citep{wei2022chain, wang2022self, zhou2022least, creswell2022selection, wang2022rationale,liu-etal-2022-multi,kojima2022large}. However, a significant distinction lies in the number of reasoning steps, which is limited and fixed in usual CoT applications. In contrast, the algorithms explored here require an unspecified number of iterations required to complete the execution, and may even incorporate double loops. The prompt contains the condition for declaring the end of execution. Interestingly, we even found that if that declaration involves the word \verb=END=, the text versions of GPT models will often stop after they produce that token, while the code-tuned versions tend to continue on, inventing a new problem to solve and solving it until they run out of tokens; Including \verb=END= as the stop word in calls to the GPT API will terminate both.

% \subsection{Application of IRSA to reasoning over logical puzzles}
\subsection{Applying IRSA to reason over logical puzzles}
\label{sec:LD}
We focus mostly on standard algorithms in order to evaluate how close GPT-3 is to being Turing-complete, but in this section, we pay special attention to a task that involves reasoning, rather than program execution. This type of task is commonly used in the NLP community for testing LLMs \cite{srivastava2022beyond}, where both natural language understanding and logical deduction are needed. The BIG-bench Logical Deduction task requires inference of the ordering of several objects given their pairwise relationships described in natural language (e.g., a robin is standing on a branch to the right of a raven, but a sparrow is the left-most). Despite the low number of objects (e.g., five) in these puzzles, LLMs struggle to solve them in zero- or few-shot settings, much like how human solvers typically cannot just see the correct answer instantly, and instead require pencil and paper to manage a search strategy. 

An LLM capable of solving such problems with only a few examples -- or even a prompt that breaks down the strategy into necessary steps in the Chain-of-Thought style -- would indeed demonstrate savant qualities. (As of the date of this writing, the logical deduction task for five objects is not solved by LLMs without external search/reasoning/inference algorithms, such as ThinkSum \cite{thinksum}). However, Prompt \ref{pr:LD} can be used to solve $76\%$ of these puzzles. The first part of this prompt has a CoT structure that translates the problem into a canonical form, isolates object names, provides the number of objects, and enumerates statements about object pairs. Then a scoring mechanism is decided on, and object scores are assigned variable names. This procedure could continue till the problem is written in an integer programming or constraint satisfaction form and then sent to an external mechanism for resolution. In \cite{thinksum}, a similar strategy was pursued to parse the puzzles into inequalities that an LLM can evaluate: Through new calls, token probabilities are used to compute the likelihood of a certain item order, so that the external reasoning step is reduced to probabilistic inference over all possible orders. 

IRSA-inducing Prompt \ref{pr:LD}, however, does not rely on external mechanisms. Instead, it provides parsing instructions based on GPT-3's English understanding, and describes iterative reasoning in a highly structured manner (similar to Prompt \ref{pr:BS}). This approach expects that the self-attention to the generated text in the response will be drawn to the strong patterns in the prompt, enabling it to continue as many iterations as needed to discover scores for the items that satisfy the ordering clues. Thus the prompt combines natural language instructions and a description of an execution path of a constraint satisfaction algorithm. (In fact, the algorithm is buggy, as discussed in Section \ref{sec:results_basic}, but still works most of the time). 

\begin{prompt*}
    \caption{A prompt for solving logical deduction puzzles from BIG-bench. \href{https://platform.openai.com/playground/p/xAOcPKewG2yK4bEwQbz8lzuj?model=code-davinci-002}{Playground link (use with 0 temperature)}}
    \label{pr:LD}
    \begin{scriptsize}
        \begin{verbatim}
PUZZLE: The following objects need to be ordered. obj1 is the biggest. obj2 is smaller than obj3. 
obj1 is bigger than obj2.
QUESTION: Which object is the biggest?

START
Parsing step:
    Items: obj1, obj2, obj3
    Numbe of items: 3
    Statement 1: obj1 is the biggest.
    Statement 2: obj2 is smaller than obj3.
    Statement 3: obj1 is bigger than obj2.
Scoring identification step:
     Scores will refer to size. 
     Since we have 3 items, let's assume that the biggest gets a score  of 3 pounds
     and the smallest gets the score of 1 pound.
Translation step:
    Available variable names: x, y, z, a, b, c
    Map item scores of 'obj1', 'obj2', 'obj3' to variable names x, y, z
    obj1 score is x; obj2 score is y; obj3 is z;
    Statement 1: 'x' is the biggest.
    Statement 2: 'y' is smaller than 'z'.
    Statement 3:  'x' is bigger than 'y'.
Initialization step:
    Words used to qualify the realtionsips: smaller, bigger, biggest
    Orientation step: 
        the biggest: refers to the score of 3
        smaller: refers to smaller score
        bigger:  refers to larger score
    Initialize so that all scores are  different numbers between 1 and 3
    Score_assignment_A:
    x=2, y=3, z=1
       
Iterative reasoning
Iteration 1:
    update_flag=false
    Statement 1:  'x' is the biggest, meaning: x should be 3
    In Score_assignment_A, x is 2
    x is not what it should be, so we need to make a change, so we set update_flag=true and we need to make a swap.
    In the statement there is only one variable and it is  x. We need  to find another. We want x to be 3,
    but we see that in Score_assignment_A that 3 is assigned to y, so we swap  values of x and y to make
    Score_assignment_B:
    x=3, y=2, z=1 
    Statement 2: 'y' is smaller than 'z', meaning: y<z
    In Score_assignment_B, y is 2 and z is 1,  so y<z maps to 2<1
    2<1 is false, so we need to make a change, so we set update_flag=true and we  need  ot make a swap.
    In the statement there are two variables and those are y and z so we swap in Score_assignment_B to make
    Score_assignment_C:
    x=3, y=1, z=2  
    Statement 3: ' x' is bigger than 'y', meaning x>y
    In Score_assignment_C, x is 3 and y is 1,  so x>y maps to 3>1
    3>1 is true, so we don't need to make a change.
End of iteration. Since update_flag is true, we need more iterations.
Iteration 2:
    update_flag=false
    Statement 1:  'x' is the biggest, meaning: x=3
    In Score_assignment_C, x is 3,  so x=3 maps to 3=3
    3=3  is true, so we don't need to make a change.
    Statement 2: 'y' is smaller than z, meaning: y<z
    In Score_assignment_C, y is 1 and z is 2, so y<z maps to 1<2
    1<2 is true, so we don't need to make a change.
    Statement 3: 'x' is bigger than y, meaning x>y
    In Score_assignment_C, x is 3 and y is 1,  so x>y maps to 3>1
    3>1 is true, so we don't need to make a change.
End of iteration. Since update_flag is false, we have finished all iterations and found the correct order.

The correct score assignment is the last (Score_assignment_C):
x=3, y=1, z=2
Reverse translation step: 
Map items 'obj1', 'obj2', 'obj3' to variable names x, y, z
so we replace x by obj1, y by obj2, and z by obj3 to get size scores:
obj1 has the score 3; obj2 has the score 1; obj3 has the score 2

Question: Which object is the biggest?
Answer: obj1
Sorting all by score starting with obj1:
with score 3, obj1
with score 2, obj3
with score 1, obj2
END

PUZZLE: On a shelf, there are five books: a gray book, a red book, a purple book, a blue book, and a black book. 
The red book is to the right of the gray book. The black book is to the left of the blue book. 
The blue book is to the left of the gray book. The purple book is the second from the right.
QUESTION: Which is leftmost?
START\end{verbatim}
    \end{scriptsize}
\end{prompt*}

\begin{prompt*}
    \caption{{\bf Fragmented prompt}: The first few Bubble Sort state transitions follow the execution path for one problem, but the path is not completed. Instead, state transitions involving \emph{different} sequences at \emph{different} execution points are provided. Initial part of the response is marked green. {\bf Skip attention}: The part of the response up to the last state is not needed to continue the generation. Only the prompt, the last state bracketed by \texttt{<state>} and \texttt{</state>}, and the text after it are necessary to generate the next token. Both single execution path prompts and fragmented prompts can have self-attention instructed to always skip to last generated state, as marked with {\bf Attend}. \href{https://platform.openai.com/playground/p/R10IV9sCAfyU1D4dpiwZys4b?model=code-davinci-002}{Playground link (use with 0 temperature)}}
    \label{pr:frag}
    \begin{scriptsize}
    \begin{curlybracketwhite*}{}{}
        \begin{verbatim}
Problem: 2, 3, 1, 5
EXECUTION
    Length of the list: L=4
    Number of pairs: P=3
    a=[2 3 1 5]
    set n_swaps=0. set i=P=3. set swap_flag=true.   
       <state> a=[2 3 1 5] i=3 P=3 n_swaps=0 swap_flag=true </state>
    Since i=3 and P=3, i and P are equal, so this iteration is done, but swap_flag is true,
    so we need another iteration
    Iteration:
       set swap_flag=false.  set i=0. The state is:
       <state> a=[2 3 1 5] i=0 P=3 n_swaps=0 swap_flag=false </state>
       Since i=0 and P=3, these two are different, so we continue
       a[i]=a[0]=2 a[i+1]=a[1]=3
       Because 2<3 is true we keep state as is and move on by increasing i
       <state> a=[2 3 1 5] i=1 P=3 n_swaps=0 swap_flag=false </state>
       Since i=1 and P=3, these two are different, so we continue
       a[i]=a[1]=3 a[i+1]=a[2]=1
       Because 3<1 is false we set swap_flag=true,increase n_swaps by one, and in a=[2 3 1 5] swap 3 and 1,
       and increase i, and keep P as is to get
       <state> a=[2 1 3 5] i=2 P=3 n_swaps=1 swap_flag=true </state>
       Since i=2 and

       <state> a=[6 5 8 9 1 2] i=2 P=5 n_swaps=5 swap_flag=false </state>
       Since i=2 and P=5 i and P are different, so we continue
       a[i]=a[2]=8 a[i+1]=a[3]=9
       Because 8<9 is true we we keep state as is and move on by increasing i
       <state> a=[6 5 8 9 1 2] i=3 P=5 n_swaps=5 swap_flag=false </state>

       <state> a=[9 1] i=0 P=1 n_swaps=2 swap_flag=true  </state>
       Since i=0 and P=1 i and P are different, so we continue
       a[i]=a[0]=9 a[i+1]=a[1]=1
       Because 9<1 is false we set swap_flag=true,increase n_swaps by one, and in a=[9 1] swap 9 and 1 
       and increase i, and keep P as is to get
       <state> a=[1 9] i=1 P=1 n_swaps=3 swap_flag=true  </state>

       <state> a=[6 7 3 5] i=3 P=3 n_swaps=7 swap_flag=false </state>
       Since i=3 and P=3 i and P are equal, so this iteration is done, swap_flag is false, so stop the iteration
Final List: 6, 7, 3, 5
Number of swaps: 7
END OF EXECUTION

       <state> a=[3 5 6 8] i=3 P=3 n_swaps=1 swap_flag=true  </state>
       Since i=3 and P=3 i and P are equal, so this iteration is done, but swap_flag is true, 
       so we need another iteration
    Iteration:
       sset swap_flag=false.  set i=0. The state is:
       <state> a=[3 5 6 8] i=0 P=3 n_swaps=1 swap_flag=false </state>

       <state> a=[2 8 1 3 5 7 4] i=1 P=6 n_swaps=5 swap_flag=false </state>
       Since i=1 and P=6 i and P are different, so we continue
       a[i]=a[1]=8 a[i+1]=a[2]=1
       Because 8<1 is false we set swap_flag=true,increase n_swaps by one, and in a=[2 8 1 3 5 7 4] swap 8 and 1
       and increase i, and keep P as is to get
       <state> a=[2 1 8 3 5 7 4] i=2 P=6 n_swaps=6 swap_flag=true  </state>

       <state> a=[4 8] i=0 P=1 n_swaps=7 swap_flag=true  </state>
       Since i=0 and P=1 i and P are different, so we continue
       a[i]=a[0]=4 a[i+1]=a[1]=8
       Because 4<8 is true we we keep state as is and move on by increasing i
       <state> a=[4 8] i=1 P=1 n_swaps=7 swap_flag=true  </state>

Problem: 3, 1, 8, 9, 6
EXECUTION
        \end{verbatim}
        \end{curlybracketwhite*}
  \definecolor{shadecolor}{rgb}{.835, .957, .824}
      \vspace{-0.8cm}

%     \begin{shaded}
%     \color{gray}
%     \begin{Verbatim}[commandchars=\\\{\}]
%     Length of the list: L=5
%     Number of pairs: P=4
%     a=[3 1 8 9 6]
%     set n_swaps=0. set i=P=4. set swap_flag=true.   
%         <state> a=[3 1 8 9 6] i=4 P=4 n_swaps=0 swap_flag=true </state>
%         Since i=4 and P=4 i and P are equal, so this iteration is done, but swap_flag is true,
%         so we need another iteration
%     Iteration:
%         set swap_flag=false.  set i=0. The state is:  
% \textcolor{black}{        <state> a=[3 1 8 9 6] i=0 P=4 n_swaps=0 swap_flag=false </state>}
% \textcolor{black}{        Since i=}
%     \end{Verbatim}
%     \end{shaded}

    % \begin{shaded}
  \begin{curlybracketgrey*}{}{}
    % \color{gray}
    \begin{Verbatim}[commandchars=\\\{\}]
    Length of the list: L=5
    Number of pairs: P=4
    a=[3 1 8 9 6]
    set n_swaps=0. set i=P=4. set swap_flag=true.   
        <state> a=[3 1 8 9 6] i=4 P=4 n_swaps=0 swap_flag=true </state>
        Since i=4 and P=4 i and P are equal, so this iteration is done, but swap_flag is true,
        so we need another iteration
    Iteration:
        set swap_flag=false.  set i=0. The state is:  
    \end{Verbatim}
    \end{curlybracketgrey*}
    % \end{shaded}
\vspace{-10pt}
  \begin{curlybracketgreen*}{}{}
    % \begin{shaded}
    \begin{Verbatim}[commandchars=\\\{\}]
        <state> a=[3 1 8 9 6] i=0 P=4 n_swaps=0 swap_flag=false </state>
        Since i=
    \end{Verbatim}
    % \end{shaded}
\end{curlybracketgreen*}

    \end{scriptsize}
    \vspace{-0.2cm}
\end{prompt*}

\subsection{Fragmented prompting}
\label{sec:fragments}
In this section, we introduce an alternative way to trigger the iterations through fragmented prompting. An example is given in Prompt \ref{pr:frag}, which differs significantly from Prompt \ref{pr:BS} in how it enables iterations of the Bubble Sort algorithm:
\vspace{-10pt}
\begin{itemize}[leftmargin=*, labelsep=1mm]
    \item {\bf Complete state specification}. In contrast to Prompt \ref{pr:BS} where iterative behaviour is induced indirectly through worked-out examples of multiple full loops, Prompt \ref{pr:frag} explicitly defines the state-to-state transitions that cause new iterations for different cases. For that to be possible, the state in Prompt \ref{pr:frag} includes the iterator $i$.
    \item {\bf Fragmentation.} Prompt \ref{pr:frag} does not fully cover the entire execution path of any single example. Instead, it follows the first three state changes\footnote{The full execution path in this style is shown in Prompt \ref{pr:BS2}.} for the sequence $2,3,1,5$, and then stops in the middle of a sentence. Then it shows 6 additional fragments of execution paths for \emph{different} problems. Each fragment illustrates a single state change.
\end{itemize}
\vspace{-5pt}

Interestingly, fragmented prompting can also trigger iterative behaviour, where the language model accurately executes the algorithm on a given input and outputs \verb=END OF EXECUTION= when the termination condition (no new updates on the sequence) is met. Viewing this prompt as an instance of in-context learning, it is challenging to classify it in usual terms. It goes beyond 0-shot learning as it contains explanations specific to the algorithmic sorting task. Yet, as opposed to what the few-shot CoT prompting might do, it does not work out any single example of array sorting. Instead, it provides fragments of patterns that can be stitched together to execute the algorithm (and GPT-3 \textsc{code-davinci-002} does execute it correctly for new inputs). 

The potential advantage of such fragmented prompting is that the prompt can be shorter and include a greater variety of situations that may be encountered in new problems without going through multiple examples (or, as in this case, without going through any single example of an execution path from start to finish). A potential disadvantage is that the language model may get confused by the fragmentation and start hallucinating independent fragments itself. In this case, we managed to avoid that by having the first fragment starting from the start of execution, going through several state transitions, and ending mid-sentence. Because of this, when a new problem is given, the language model starts running the execution path from the beginning, and later refers to various cases in the prompt for guidance on how to proceed. The \verb=<state>*</state>= structure encourages the model to refer to the previous state it just generated when creating the follow-up text.

\subsection{Skip attention}
\label{sec:skip}
Prompt \ref{pr:frag} also illustrates the idea of attention skipping. Whether we are using a fragmented prompt or a single-execution prompt, if the state in the \verb=<state>*</state>= structure is complete, the attention mechanism can effectively generate the next token without attending to all the generated text. It is sufficient to attend to the prompt and the text generated after and including the last state. 

Depending on the implementation, such skipping can offer several advantages. If the skipping is implemented on the server side, akin to how OpenAI provides stop word functionality in its API, then skipping unnecessary attention saves computation: The state of the model at the end of the prompt is cached and used to continue processing from the latest generated \verb=<state>= marker, ignoring the text generated in-between. Skip-to-state can also be implemented on the client side, iteratively updating the original prompt by concatenating the latest \verb=<state>*</state>= structure to the original prompt and calling the generative model with \verb=</state>= as a stop sequence (We did the latter in our experiments). In both cases, the skip-to-state strategy should increase the number of tokens that can be generated, as self-attention, which grows linearly with the generated text, is the primary cause for the token limitations. Skip-to-state strategy keeps the self-attention cost constant. As IRSA requires the unrolling of potentially long iterative algorithms, these savings are important. For example, running a dynamic program that keeps track of 2D matrices is only practical in this manner. (See also \cite{schuurmans2023memory} on an external memory approach to dealing with limited attention length. Here we deal with it by skipping parts of generated text, instead).
Another advantage of skip-to-state attention is that by only attending to the necessary information, the generative model is less likely to get confused by accidental patterns created in its own generated text. (See more on this in Section \ref{sec:discussion} and Figure \ref{fig:long_pattern}.)

\section{GPT as a machine language: Prompting to interpret/compile a program}
\label{sec:interpret}

Previous examples show that GPT-3 can be prompted to perform iterative reasoning and program execution by providing it with execution paths or multiple path fragments, which implies that GPT-3 is close to being Turing-complete. On the other hand, a general-purpose computer can also execute algorithms that convert the text of a program into its machine code. Indeed, we can design prompts with instructions on how to turn code in some language into execution paths that can then be used in prompting. 

An example is shown in Prompt \ref{pr:interpret}, where several examples of hypothetical syntax for transforming states are given, including setting values of variables and matrices, printing them, a single loop program execution, and the \verb=detailed_max= function that breaks down steps and explains them. Then, the double loop dynamic programming algorithm for finding the longest common subsequence (LCS) is also presented in this new language. This prompt successfully triggers the correct execution of the algorithm, complete with detailed explanations and state transitions (green shaded in Prompt \ref{pr:respLCS}). This can then be used as a prompt to execute the LCS algorithm on arbitrary inputs (Section \ref{sec:results_skip}). We should note that GPT-3 is still sensitive to small alterations in text, and Prompt \ref{pr:interpret} does not always lead to good interpretations of the algorithm. The performance may depend on accidental deceptive patterns and inconsistencies in the prompt, as well as the input. Nevertheless, once the output has been verified as correct, the Prompt \ref{pr:interpret} together with the response in Prompt \ref{pr:respLCS} became the prompt -- IRSA 'machine code' for GPT --- to execute (mostly correctly) the LCS algorithm for new inputs, as long as they are appended in the same format:
\vspace{-5pt}
\begin{verbatim}
LCS:
Input: <seq1> <seq2> End of input
LCS Prep:
\end{verbatim}
\cut{
The stop sequence \verb=END= is used to determine when to stop generation, as otherwise, the model may finish with the given input, then imagine a new one -- or a new task all together -- and try solving it.}
\vspace{-10pt}

\section{Results}
\label{sec:results}

We evaluated two versions of iteration by regimenting self attention (IRSA):
\vspace{-10pt}
\begin{itemize}[leftmargin=*, labelsep=1mm]
    \item {\bf Basic IRSA}: Prompting with highly structured single execution path examples (Table \ref{tab:main}). Although similar to CoT prompting, there are notable differences. CoT prompts typically provide multiple steps of reasoning shown for a few examples and have the LLM perform the same steps on a new example. Conversely, IRSA prompts are designed to trigger iterative reasoning that is repeated until the stop condition is reached and the solution is found. Furthermore, the execution path example for each task is deliberately chosen to be out-of-distribution (e.g., the Bubble Sort prompt features a worked-out example of sorting a four-number sequence in just three passes, while the dataset consists of five-number sequences requiring 2 to 5 iterations and up to 20 state transitions, with varying complexity across problems). Thus in terms of information they provide, these prompts can be seen as somewhere between single-shot and zero-shot prompts.
    \item {\bf Skip-to-state IRSA}: Prompting as above, but with additional forced attention skipping. In this approach, the LLM is forced to attend only to the prompt and the last generated state as it iterates through the input to find the solution (as illustrated at the end of Prompt \ref{pr:frag}). Our experiments (Table \ref{tab:skip}) also evaluated fragmented prompts, where the prompt does not consist of a single complete execution path for an example, but instead shows several state-to-state transitions for different inputs. Such prompts help generalize the instructions, even though they further deviate from the single-shot prompts, as there is no single example worked out till the end. 
    \end{itemize}
\vspace{-10pt}

{\bf Baselines.} To make fair comparisons and avoid unnecessary recomputation, we reused existing baselines from \citet{srivastava2022beyond} wherever possible, denoted by an asterisk (*) (especially considering that these baselines typically perform close to random guessing on certain tasks).
% in light of the large difference in performance between our approach and the baselines, which on these tasks are typically very close to random guessing). 
We reused these datasets and baselines for the following tasks: Logical deduction, Balanced parenthesis, and Longest common subsequences for long sequences. We created our own datasets and ran baselines for the following tasks: Bubble sort, Longest substring without repeating characters, and Longest common subsequence for short sequences. We include the best result from \citet{srivastava2022beyond} for the GPT family, as our experiments were mainly conducted using GPT-3. Our baselines included zero or few (5) shot prompting with or without relevant code added to the description of the task in the prompt (e.g. Prompt \ref{pr:FewShotsCode}). Few shot baselines were made with 5 different random choices of examples to be included in the prompt. 

Table \ref{tab:main} summarizes the success of the basic IRSA in comparison to the best-performing baselines. Table \ref{tab:skip} provides a breakdown of individual baseline performance and evaluates the IRSA using skip attention and fragmented prompting. The 'Guessing' strategy refers to picking the most frequently correct answer for a given task as a guess for each problem in the task, which is different from truly random guessing. In-context learning with multiple examples could prime the answers to pick the most frequently seen answer, even when no understanding of the problem happens. Thus, our 'Guessing' strategy captures the task difficulty more accurately.

{\bf Models.} We have briefly experimented with different members of the GPT-3 family, but decided to run complete experiments with \textsc{code-davinci-002} for two reasons: \textsc{text-davinici-002} and \textsc{003} often produced qualitatively similar results, and experimentation with the lightweight \textsc{code-davinci-002} was easier due to better combination of token quota and availability. Having been tuned on code, this model may have slight advantages over models tuned for other more natural language-targeted tasks. Nevertheless, as we show in the experiments and discuss in Section \ref{sec:discussion}, without IRSA, \textsc{code-davinci-002} cannot solve the problems we discuss here, even when it can generate the code that could. To induce iterative reasoning in LLMs, it appears that attention needs to be highly regimented through strong structure, and possibly additional attention control, such as the skip-to-state strategy we described in Section \ref{sec:skip}. This also applies to GPT-4~\cite{openai2023gpt4}, which, at the time of this writing, has just become available for testing. In the Discussion (Section \ref{sec:gpt-4}), we show that prompting GPT-4 with straight-forward Prompts \ref{pr:GPT4savant}, \ref{pr:GPT4steps1}, \ref{pr:GPT4steps2} does not match the performance of IRSA in GPT-3.

\subsection{Datasets}
{\bf Bubble sort.} We created a dataset of 100 random non-repeating digit sequences of length 5. For each sequence, we ran the bubble sort algorithm to establish the total number of element swaps it requires. The task is to predict the number of swaps for a given sequence.

{\bf Longest substring without repeating characters.} A classical coding interview question: Given a string of letters, find the longest contiguous substring such that no letter appears more than once. We created a dataset of 100 random strings of length 7, and for each found the length of the longest subsequence without repeating characters. The task is to predict that length for a given sequence.

{\bf Logical deduction \cite{srivastava2022beyond}.} We include this task (Section \ref{sec:LD}) in experiments to emphasize the broad importance of triggering iteration in LLMs responses. Enabling LLMs to execute iterative algorithms through effective prompting could help solve numerous reasoning problems. To illustrate this, we consider this task that involves solving a puzzle about an order of items/objects/persons, such as books on the shelf, birds on a branch, cars, golfers, etc., given several clues, such as ``minivan is more expensive than the car'', or ``the robin is to the left of the finch.'' We focus on a subtask involving 5 items, with varying sets of items and the types of ordering across the puzzles. While in-context learning with LLMs consistently solves less than $35\%$ of puzzles, a recent combination of GPT-3 and probabilistic reasoning \cite{thinksum} was able to solve $77\%$ of the puzzles. We reach a similar performance through IRSA, \emph{without} an additional external reasoning mechanism.

{\bf Valid parentheses \cite{srivastava2022beyond}.} The task is the first of the two in the cs-algorithms challenge in BIG-bench. The goal is to evaluate LLMs ability to perform reasoning equivalent to the classical stack manipulations needed to verify if a sequence of parentheses of different types is balanced or not. LLMs (including GPT) tend to do about the same as chance ($50\%$), except for PaLM with 3 shots, which gets around $75\%$ accuracy.

{\bf Longest common subsequence (long) \cite{srivastava2022beyond}.} The second task in BIG-bench cs-algorithms involves solving the classical dynamic programming problem. Defining a subsequence of a sequence to be a sequence of symbols one could get by skipping arbitrary stretches in the original sequence, the task is to find the length of the longest subsequence common to two given sequences. LLMs do not do much better than chance on this task ($\sim$10\%).

{\bf Longest common subsequence (short).} We created this dataset in the same manner as the above one from the BIG-bench, but with the constraint on the sequence lengths, limiting them to a maximum of 6 characters. This allows us to evaluate IRSA on more cases, ensuring it does not run out-of-memory (tokens) in generation\footnote{Buble sort, Longest substring, and Longest common subsequence (short) datasets: \href{https://github.com/anajojic/gpt-coding}{https://github.com/anajojic/gpt-coding}}.

\begin{table}[t]
\small
\centering
\def\arraystretch{1.15}
\begin{tabular}
{l|ccc}
\toprule
\textbf{Task} & \textbf{IRSA} & \textbf{Baseline} & \textbf{Guessing} \\
\midrule
Bubble sort &  & &  \\
\quad - Prompt \ref{pr:BS} & $0.74$ & $0.27$ & $0.23$ \\
\quad - Prompt \ref{pr:BS2} & $1.00$ & $0.27$ & $0.23$ \\
Longest substring & $1.00$ & $0.60$ & $0.59$  \\
Logical deduction & $0.76$ & $0.32^*$ & $0.2$  \\
Parentheses & $0.96$ & $0.56^*$ & $0.5$ \\
%\multirow{2}{*}{\parbox{2.3cm}{Longest common subseq. (short)}} & $\mathbf{?_{\pm 0.86}}$ & $\mathbf{?_{\pm 0.30}}$ & $0.44$  \\
%& & & \\
\bottomrule
\end{tabular}
% \vspace{-5pt}
\caption{ Iteration through Regimented Self-Attention (IRSA) compared with standard in-context learning baselines, and with the strategy of always guessing the most frequent answer. (*) denotes the best result for GPT-3 from the BIG-bench~\cite{srivastava2022beyond}
% \href{https://github.com/google/BIG-bench}{BIG-bench Github}
.}
\label{tab:main}
\vspace{-10pt}
\end{table}
\subsection{Basic IRSA results}
\label{sec:results_basic}

The basic IRSA involves a single prompt for a given category of problems (or algorithms). The prompt contains a single execution path for one problem, deliberately to be slightly out of the data distribution, such as a shorter problem than those found in the datasets. The results are summarized in Table \ref{tab:main}. For Bubble Sort evaluations, we show the results using both Prompt \ref{pr:BS}, and Prompt \ref{pr:BS2}. The latter is a single execution path for the same problem ($2,3,1,5$), but in the style of Fragmented Prompt \ref{pr:frag} by continuing the execution path initiated by Prompt \ref{pr:frag}, without incorporating fragments from other paths. The former had an accuracy of $74\%$ for inferring the numbers of swaps necessary to sort different sequences, while the latter achieved $100\%$. Note that while the execution path for the example $2,3,1,5$ requires three iterations of the outer loop and three iterations in each inner loop, the dataset contains sequences of length 5 and thus requires four iterations in the inner loop and a variable number of iterations of the outside loop -- anywhere from 2 to 5 -- and yet the model can execute the correct number of iterations based on the stoppage criterion (that in the inner loop, no changes were made to the sequence).

For the logical deduction puzzles, we used the Prompt \ref{pr:LD}. Note that the logic of the iterative reasoning there is faulty as it may enter an infinite loop. When that happens, the generation runs out of tokens and we simply used the answer after the 4th iteration in evaluation. Further discussion in Section \ref{sec:discussion} suggests the potential for creating more effective prompts. Nevertheless, with this prompt to induce iterative reasoning, we still reach the state-of-the-art results, comparable only with \cite{thinksum}, which uses an external reasoning mechanism in conjunction with prompting.

To solve the longest substring without repeating characters problems, we developed Prompt \ref{pr:lss} based on the 1-index version of the following single-pass algorithm. Interestingly, this algorithm trades computation for memory by creating one variable per unique letter in the sequence for storing the location where the letter was last seen in the sequence during the pass (\verb=last_ind=):
\vspace{-5pt}
\begin{verbatim}
# s contains the given string
last_ind = {}
m_len = 0 

# window start
st_ind = 0
 
for i in range(0, len(s)):
  if s[i] in last_ind:
    st_ind=max(st_ind,last_ind[s[i]]+1)
 
  # Update result if window is longer
  m_len = max(m_len, i-st_ind + 1)
 
  # Update last index of the character
  last_ind[s[i]] = i 
return m_len
\end{verbatim}
\vspace{-5pt}

To address the parentheses problem, we used the single execution path that demonstrates the needed stack operations for determining whether the sequence is balanced or not. The beginning and the end are shown in Prompt \ref{pr:para}. For brevity, we have omitted certain portions represented by ellipses. Note that creating long prompts is made easier by GPT's autocompletion capabilities, i.e., by starting with a description of a few steps and asking the model to complete it. Then wherever we want the prompt to differ from the model's guess, we erase the generated text from that point and continue typing our correction/instruction and try to autocomplete again. (See also Section \ref{sec:discussion}, as well as Section \ref{sec:interpret}). But interestingly, as discussed in Section \ref{sec:fragments} on fragmented prompting, parts of the execution paths can be omitted: In fact, the Prompt \ref{pr:para} as is, with the ellipsis instead of 10 steps in the algorithm, still achieves $91\%$ accuracy!

\begin{table}
\small
\centering
\def\arraystretch{1.15}
\begin{tabular}
{l|cccc}
\toprule
\textbf{Classical prompting} & \textbf{Bubble sort} & \textbf{LCS-S} & \textbf{LCS-L} \\
\midrule
0-shot & $0.20$ & $0.09$ & $0.14^*$  \\
0-shot + code & $0.20$ & $0.11$ & -  \\
few shot & $0.25_{\pm 0.05}$ & $0.07_{\pm 0.01}$ & $0.16^*$  \\
few shot + code & $0.23_{\pm 0.03}$ & $0.06_{\pm 0.02}$ & -  \\
Guessing & $0.23$ & $0.44$ & $0.10$ \\
\midrule
\textbf{IRSA skip-to-state} &  &  & \\
\midrule
single path & $0.95$ & $0.93$ & $0.28$  \\
7 fragments & $0.99_{\pm 0.02}$ & - & -  \\
13 fragments & $0.97_{\pm 0.03}$ & - & -  \\
19 fragments & $0.99_{\pm 0.02}$ & - & -  \\
25 fragments & $0.97_{\pm 0.03}$ & - & -  \\
%& & & \\
\bottomrule
\end{tabular}
% \vspace{-5pt}
\caption{ IRSA with forced skipping of everything but the latest state in generated text for Bubble Sort and Longest Common Subsequence. We also evaluate fragmented prompting in Bubble sort, where prompts contain multiple state-to-state fragments of execution paths for different problems. (*) denotes the best result for the GPT family from the BIG-bench~\cite{srivastava2022beyond}} 
\label{tab:skip}
\vspace{-10pt}
\end{table}

\subsection{Skip-to-state attention results}
\label{sec:results_skip}
The dynamic programming solution to the longest common subsequence (LCS) problem involves a large state that includes a $(M+1) \times (N+1)$ matrix representing the solution for all prefixes of the two sequences of lengths $M$ and $N$. Without skip-to-state attention (Section \ref{sec:skip}), the API calls run out of tokens before reaching the end for all but the shortest problems. As elaborated in Section \ref{sec:interpret} we used Prompt \ref{pr:interpret} to compile an execution path in Prompt \ref{pr:respLCS}, and then used both of them to induce IRSA on LCS short (LCS-S) and LCS long (LCS-L) problems. Even with skip attention, the state was too large to fit the token limit for most of the problems in LCS-L from BIG-bench. Yet, IRSA with skip attention still beats the state-of-the-art significantly (Table~\ref{tab:skip}). On shorter problems LCS-S, where the state was small enough to allow IRSA with skip attention to make all state-to-state transition without running out of tokens, the performance was a respectable $93\%$. Note that even GPT-4, without IRSA, cannot reach this accuracy (Section \ref{sec:gpt-4}).

We also tested fragmented prompting of Bubble Sort execution, as shown in Table~\ref{tab:skip}. Interestingly, state-to-state skipping slightly hurts the performance of the single execution path prompt that achieved $100\%$ accuracy when used without skipping (Table \ref{tab:main}). Nonetheless, the fragmented prompts did well with skip attention. For each selected number of fragments -- 7, 13, 19, 25 -- at least one of five randomly generated prompts achieved $100\%$ accuracy. These prompts followed the format in Prompt \ref{pr:frag}, starting with the few state transitions from the beginning for the sequence $[2,3,1,5]$ and then listing additional 6, 12, 18, or 24 fragments. Bubble Sort has 6 different transitions, and listing one, two, three, or four of each type, with a random sequence in the state, allows fully balanced instruction, which leads to a slightly better performance than having a completely randomly chosen execution path fragment (data not shown). These six basic transitions, illustrated in Prompt \ref{pr:frag}, involve two ways of ending an iteration depending on the swap flag and four ways of changing the state: two possibilities for inequality being true or not, combined with two possible previous values of the swap flag. As we could have expected, using an ensemble approach with multiple prompts can yield $100\%$ accuracy, even if the individual prompts fall short: The cryptic sensitivity to accidentally hidden (at least to our eyes) patterns in the prompts means that different prompts fail for different test cases. Thus, every one of the fragmented prompt collections yields $100\%$ when used as an ensemble.

\section{Discussion}
\label{sec:discussion}

Iteration by Regimenting Self-Attention (IRSA) is a technique for \emph{triggering code execution in GPT-3 models}. Note that the goal is different from the goal of Alphacode \cite{alphacode} and Copilot \cite{copilot, copilot_study}, which are meant to \emph{write} the code, without necessarily understanding what it outputs. While there are indeed examples of rather impressive code generation and even, anecdotally, execution path generation using minimal prompting in the latest Codex and GPT-3 models, the lack of control in current LLMs prevents the consistent achievement of these feats with precision, 
% However, without much more control current LLMs cannot consistently pull off these feats or with any accuracy, 
which is why the code generation applications involve humans in the loop. For instance, as illustrated in zero-shot bubble sort code Prompt \ref{pr:zeroBBScode}, when relying on Codex alone to attempt code execution, the generated samples are intuitively close to the correct solution, but a bit off, preventing correct execution. IRSA, on the other hand, can produce consistently accurate outputs.

In algorithm design, trading computation for memory use is a recurrent idea. IRSA as a technique for LLM inference can be seen in a similar light: We could train a bigger model on more data, with attention spanning deeper into the past tokens, hoping that it could answer a simple yet computationally complex query in just a couple of tokens directly; or we can devise a prompting strategy instructing a smaller LLM to use its token stream as a memory tape, allowing it to reach similar functionality with increased token usage. By triggering and controlling iterative behaviour, we can, in principle, execute arbitrary algorithms, which further raises interesting questions: What are the consequences of LLMs becoming Turing-complete? And how difficult is it to program via IRSA? Will larger GPT models become capable of executing programs correctly without IRSA? Based on our experience in designing the prompts we showed here, we speculate on these three questions in this section.

\subsection{Possible consequences}
\cut{
}

{\bf (Teaching) Coding.} The integration of LLMs' code generation capabilities with IRSA leads to innovative applications in code generation. Some of it is implied in the interpreter/compiler Prompt \ref{pr:interpret}, which instructs GPT how to interpret and execute code. Following these ideas, exploring program verification and automatic debugging could be a promising direction. Another obvious application of IRSA is in computer science education, where we often expect students to execute programs on paper to determine what the state will be at some point during the execution. Furthermore, IRSA may also point to new ways of programming by example. 

{\bf Adversarial applications.}
Any time a computational medium is Turing-complete, a variety of malicious uses may become possible, such as creating and executing malware, exploiting system vulnerabilities, conducting cryptographic attacks, causing resource exhaustion, etc. Thus we should be aware of the double-edged sword with the increased versatility and computational power of GPT models.

{\bf In-context learning and LLM evaluation.}
 Prompting with IRSA must be considered a zero- or one-shot learning technique, analogous to chain-of-thought prompting. If, via IRSA, LLMs can be disciplined with a regimented prompt to execute arbitrary algorithms involving (double) loops, they may be able to solve arbitrary problems NLP researchers can compose, incorporating natural language understanding and iterative reasoning like belief propagation, constraint satisfaction, search, etc. This renders many of the hard BIG-bench tasks easier than they initially appear, as already suggested by \cite{suzgun2022challenging} using classical CoT prompting. Many CoT results can be further improved with IRSA (as logical deductions with Prompt \ref{pr:LD}). 
 
 However, triggering such iterative behaviour may still be hampered by the same sensitivity of in-context learning to accidental misleading patterns, already observed in classical prompting \citep{lu-etal-2022-fantastically,zhao2021calibrate}, where there may exist a ``fantastical'' crafting of the prompt that significantly improves the accuracy of the task. In fact, iterative reasoning may further amplify the fantastical choices. Thus, if one LLM successfully solves a hard logical reasoning task using a suitable prompt while another does not, this might imply that the optimal prompt has not yet been found. In fact, it would not be surprising if better prompts are eventually found that enable the LLM we used here (GPT-3, \textsc{code-davinci-002}) to solve all tasks with $100\%$ accuracy. Thus, evaluating LLMs on their in-context learning abilities is of questionable value: Some of the hard tasks in BIG-bench may be better suited to evaluating the skills of prompt engineers rather than the LLMs themselves.

{\bf Hybrid models -- LLMs as translators.}
If LLMs are Turing-complete and can transform problems described in natural language into algorithmically solvable programs, the decision to let them execute the program or not becomes a practical matter of computational cost. With the apparent magic of savant-like guessing gone, it is much more practical to run the algorithms on a classical computer, an approach taken by, for example, \cite{thinksum} where the external computational mechanism performs probabilistic inference, or \cite{decomp} that involves external control flows, and many other recent published and unpublished experiments combining LLMs with external calls and tools~\cite{parisi2022talm, gao2022pal, yao2022react, press2022measuring, schick2023toolformer, paranjape2023art}. Such hybrid models could separate the higher level reasoning ``System 2'' -- to use an analogy with models of human cognitive processes \citep{tversky1974judgment,kahneman} -- from the lower-level ``knee-jerk reaction'' reasoning ``System 1'', however savant-like it might be. In such systems, LLMs can dramatically improve traditional artificial intelligence algorithms simply by translating the problems into an appropriate form: see Prompt \ref{pr:wolfram} where the logical deduction task is solved by creating a call to the \verb|Solve| command in Wolfram language (Mathematica) for an example. The artificial intelligence community is increasingly interested in researching such systems, e.g., \citep{bengio1,bengio2}, and the developer community is already developing and deploying hybrid language models (Bing-ChatGPT integration, for instance).

{\bf Self-attention control in training and inference.}
To paraphrase an old adage on parenting, researchers have spent a lot of effort teaching GPTs to pay attention to everything in the text, and now IRSA is an attempt to stop it from attending to everything. We accomplish it both by drawing attention with a strong repetitive structure and by brute force through skip attention (Section \ref{sec:skip}). More flexible ways of determining what the model should attend to may be needed both in model building and inference.

\subsection{Pitfalls of programming in GPT-3}
Prompts we experimented with induce single loop or double loop program execution. Generally, controlling double loop algorithms, such as Bubble Sort and Longest Common Subsequence, is more challenging. The difficulty lies not in understanding the double loop logic, but rather in the increased probability of running into some of the problems described below. These problems are not always obvious, but can result in a wide range of accuracies achieved by seemingly similar prompts. For example, the two prompt designs for Bubble Sort both worked surprisingly well, but showed a big gap in performance between them ($74\%$ and $100\%$). Here are some tips for attempting IRSA.

{\bf Keep a complete state.} While it is often possible to instruct by analogy without fully accounting for all decisions, keeping the full state (i.e., showing it repeatedly after each transition) is usually preferable. For example, Prompt \ref{pr:frag} contains the iterator variable in the state, while Prompt \ref{pr:BS} does not. Not only does keeping full state help regiment the attention, but it makes fragmented prompting and skip-to-state attention possible.

{\bf Explain why before the instruction, not after.} LLMs are autoregressive, which makes them easier to prompt in order: from left to right. Thus, instead of instructing with `\verb|We now swap 4 and 2 because 2<4|', we instruct with:

\verb|Because 4<2 is false we swap 4 and 2|

Then later in generation, e.g., `\verb=Becasue 5<3 is=' will trigger generation of token \verb|false| and it, in turn, will trigger generation of `\verb|we swap|', and so on.

{\bf Avoid unnecessary variation, follow strong structure.} We used the term \emph{regimenting} attention in the naming of the technique to emphasize that strong structure is even more important in IRSA than in other prompting applications. It is usually crucial to order the variables in the state always in the same order, utilize the same keywords to designate the state, use the same language to explain the transitions, and ensure consistent capitalization, punctuation, and even spacing/tabulation. We experimented with several variants of the Bubble Sort prompt, and even when using the same worked-out example, the accuracy can vary dramatically.

{\bf Generate as much of the prompt with LLM itself.} One way to create such a strong structure is to let the model continue the prompt we are designing after every few lines (going back to correct the incorrectly generated continuation). The model is more likely to stay faithful to the pattern human started than the human is (with spacing, typos, and so on). Because of this, using the interpreter/compiler Prompt \ref{pr:interpret} to create an LCS execution path to serve as a prompt is a safer way of generating an IRSA-inducing prompt (as long as we verify that the exemplary execution path is correct).

{\bf Overlapping patterns can be problematic.} 
When generating the next token, an LLM has to balance many influences of patterns both in the prompt and the so-far generated text. For example, in the LCS algorithm execution Prompt \ref{pr:respLCS}, the model has to balance the long-range self-attention when deciding the next token after \verb|C[1,1]=| with the short-range influences, which make the token \verb=1= most likely after two \verb=1=s in a row regardless of the longer context. At times, short-range influences prevail and cause an incorrect execution. But, long-range self-attention can also inappropriately overrule correct short-range reasoning. For instance, when generating based on the Bubble Sort Prompt \ref{pr:frag}, the model generates repetitive text that includes many statements of the form  `\verb=Because =$n$\verb=<=$m$\verb= is true=$/$\verb=false ...=,' which can create strong pattern overruling local evaluation of the next inequality. To demonstrate that, we evaluated the likelihood of the next token after `\verb|Because 2<1 is|' for different lengths of context preceding this text. The context had between 1 and 15 lines of text in the form  `\verb|Because 2<|$m$\verb| is true we ...|' with $m\in[3..9]$ randomly chosen, e.g.
\begin{verbatim}
Because 2<3 is true we ...
Because 2<7 is true we ...
Because 2<5 is true we ...
Because 2<1 is
\end{verbatim}
\begin{figure}[t!]
    \centering
   \includegraphics[width=0.8\linewidth]{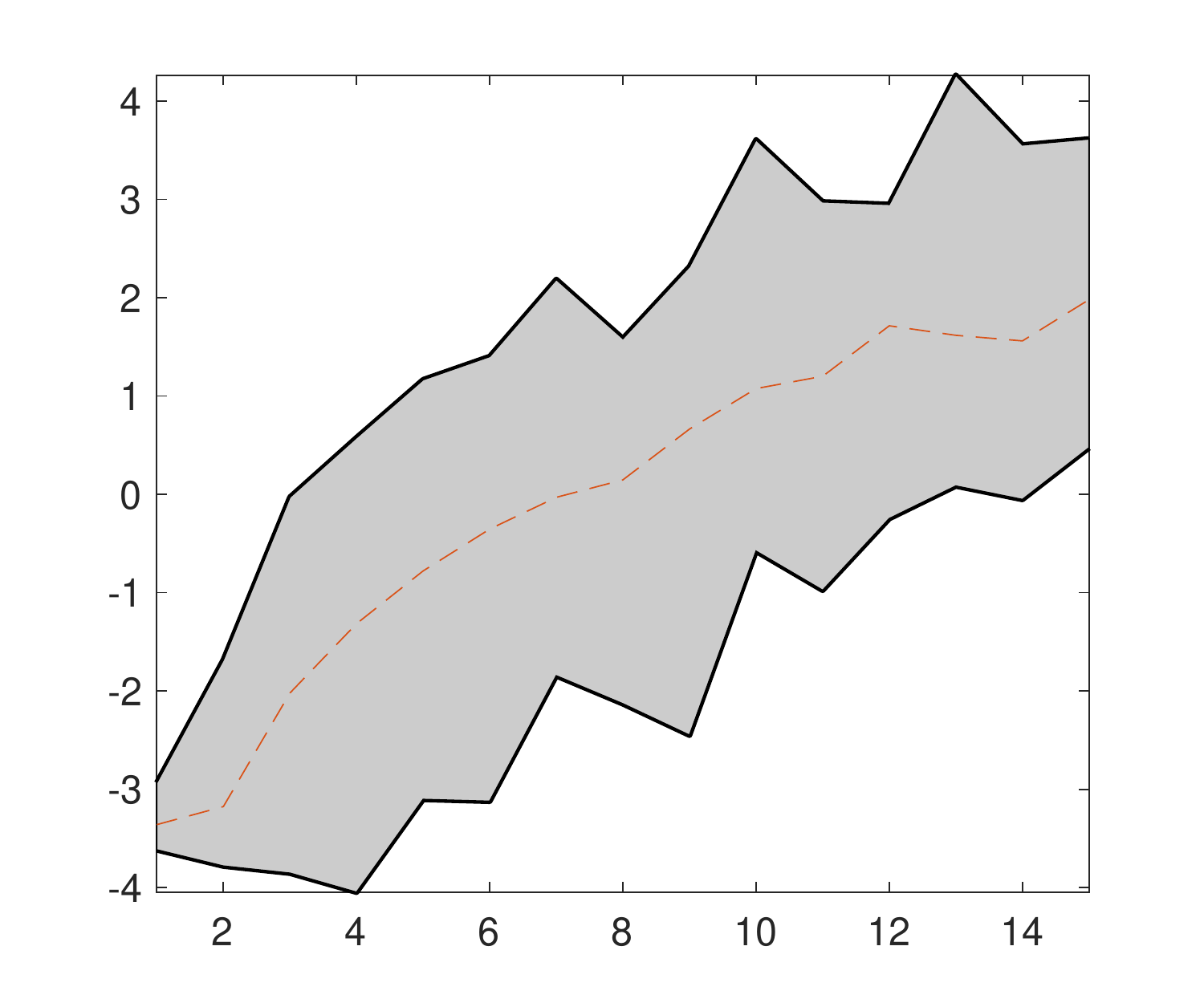}
   \vspace{-0.2cm}
   \caption{ The difference between GPT Codex log probabilities of tokens \texttt{true} and \texttt{false} after `\texttt{Because 2<1 is}', which was preceded by a long context of variable length (x-axis). The context contains between $1$ and $15$ lines of text comparing number 2 with randomly chosen \emph{larger} numbers and declaring, e.g., \texttt{Because 2<6 is true ...} We show the band between the maximum and minimum log odds over 20 trials, as well as the mean of the difference. When the preceding context does not have too many comparisons of $2$ with larger numbers, the model overwhelmingly prefers the correct evaluation \texttt{false}, but when the context is longer than 7 statements, the model usually prefers \texttt{true}. }
    \label{fig:long_pattern}
\end{figure}
As we show in Fig \ref{fig:long_pattern}, although the preceding context is correct when evaluating the inequalities, the log odds of an incorrect evaluation of \verb=2<1= increase by over six orders of magnitude with the length of this context. The longer this context is, the more it reinforces the pattern `\verb|Because 2< ... true|': If $2$ was smaller than a variety of numbers, then it is smaller than $1$, too! Furthermore, there is a large variation due to the random selection of $m$ in the examples in the context, indicating a variety of other patterns that drive the generation (The figure shows the band between the maximum and minimum log odds over 20 runs). For the contexts of length 7 the odds of picking \verb=true= over \verb=false= become roughly even. IRSA can drive probabilities to be so taut that rerunning the same API call with zero temperature can sometimes return a different result (The code behind the API presumably always adds a very small constant to log probabilities before sampling). Skip-to-state strategy in Section \ref{sec:skip} is thus less sensitive to patterns that result from program execution. 

This fragility further emphasizes the difficulty in evaluating LLMs on in-context learning tasks: Improving accuracy may simply be a matter of spending more time designing a prompt (becoming a GPT whisperer). Still, getting GPT to execute the algorithms studied here was not excessively hard, and it may even become easier on newer models.

\subsection{And what about GPT-4?}
\label{sec:gpt-4}
A recent qualitative analysis of GPT-4 abilities \cite{bubeck2023sparks} includes one example of detailed execution of a Python program for one input (in their Fig. 3.7). The LCS  algorithm is well-known, so would the newer and better GPT-4 model execute it correctly and consistently across different inputs? In Prompt \ref{pr:GPT4savant}, we show a prompt that simply asks GPT-4 to show the LCS algorithm, execute it, and report the result. On our LCS-S dataset, using this prompt design and sampling with zero temperature, GPT-4 gets the correct answer 49\% of the times, just slightly better than the 'Guessing' baseline (Table~\ref{tab:main}). An alternative prompt shown in Prompt \ref{pr:GPT4steps1}, asks for intermediate steps of execution to be shown before the answer is generated, moving the prompting strategy closer to IRSA. This prompt can be thought of as a version of Prompt \ref{pr:interpret}, but lighter and more straightforward, expecting GPT-4 to be able to show program execution without strict specifications. This prompt leads to the accuracy of 69\% on LCS-S, still behind IRSA result with codex (93\%, Table \ref{tab:skip}). To illustrate why this may be, in Prompt \ref{pr:GPT4steps2} we show the same prompt asking for intermediate steps, but for a different input. The inputs in Prompts \ref{pr:GPT4steps1} and \ref{pr:GPT4steps2} were processed differently, even though everything else in the prompts was the same, and API calls were made with zero temperature. In one case, only the initial and end states of the ``dp'' matrix are shown, while in the other, several steps (but not all!) are shown. Therefore, it seems that GPT-4 is still hard to control without regimenting self-attention more strictly.

\section{Conclusion}
We demonstrated that GPT-3 can be triggered to execute iterative algorithms, including double loops with variable termination conditions, by careful prompt design, which emphasizes structure in an example of an execution path of the desired algorithm, or in a collection of fragments of such paths. A variety of algorithms we were able to execute indicate that the GPT family is already close to being Turing-complete, i.e., not only capable of recalling or composing (hallucinating) code, but actually simulating its execution. This has several consequences. For example, if LLMs are Turing Machines (in addition to being natural language translators and analyzers), their evaluation probably needs to be rethought, esp. in cases where models are expected to make inferences for which we have algorithms, because in-context learning would cover prompts designed to execute them (Section \ref{sec:discussion}). 

\bibliography{main, anthology}

\begin{thebibliography}{42}
\providecommand{\natexlab}[1]{#1}
\providecommand{\url}[1]{\texttt{#1}}
\expandafter\ifx\csname urlstyle\endcsname\relax
  \providecommand{\doi}[1]{doi: #1}\else
  \providecommand{\doi}{doi: \begingroup \urlstyle{rm}\Url}\fi

\bibitem[Bengio(2017)]{bengio1}
Bengio, Y.
\newblock The consciousness prior.
\newblock \emph{arXiv preprint arXiv:1709.08568}, 2017.

\bibitem[Brown et~al.(2020)Brown, Mann, Ryder, Subbiah, Kaplan, Dhariwal,
  Neelakantan, Shyam, Sastry, Askell, Agarwal, Herbert-Voss, Krueger, Henighan,
  Child, Ramesh, Ziegler, Wu, Winter, Hesse, Chen, Sigler, Litwin, Gray, Chess,
  Clark, Berner, McCandlish, Radford, Sutskever, and Amodei]{GPT3}
Brown, T., Mann, B., Ryder, N., Subbiah, M., Kaplan, J.~D., Dhariwal, P.,
  Neelakantan, A., Shyam, P., Sastry, G., Askell, A., Agarwal, S.,
  Herbert-Voss, A., Krueger, G., Henighan, T., Child, R., Ramesh, A., Ziegler,
  D., Wu, J., Winter, C., Hesse, C., Chen, M., Sigler, E., Litwin, M., Gray,
  S., Chess, B., Clark, J., Berner, C., McCandlish, S., Radford, A., Sutskever,
  I., and Amodei, D.
\newblock Language models are few-shot learners.
\newblock \emph{Neural Information Processing Systems (NeurIPS)}, 2020.

\bibitem[Bubeck et~al.(2023)Bubeck, Chandrasekaran, Eldan, Gehrke, Horvitz,
  Kamar, Lee, Lee, Li, Lundberg, Nori, Palangi, Ribeiro, and
  Zhang]{bubeck2023sparks}
Bubeck, S., Chandrasekaran, V., Eldan, R., Gehrke, J., Horvitz, E., Kamar, E.,
  Lee, P., Lee, Y.~T., Li, Y., Lundberg, S., Nori, H., Palangi, H., Ribeiro,
  M.~T., and Zhang, Y.
\newblock Sparks of artificial general intelligence: Early experiments with
  gpt-4, 2023.

\bibitem[Chen et~al.(2021)Chen, Tworek, Jun, Yuan, Pinto, Kaplan, Edwards,
  Burda, Joseph, Brockman, Ray, Puri, Krueger, Petrov, Khlaaf, Sastry, Mishkin,
  Chan, Gray, Ryder, Pavlov, Power, Kaiser, Bavarian, Winter, Tillet, Such,
  Cummings, Plappert, Chantzis, Barnes, Herbert-Voss, Guss, Nichol, Paino,
  Tezak, Tang, Babuschkin, Balaji, Jain, Saunders, Hesse, Carr, Leike, Achiam,
  Misra, Morikawa, Radford, Knight, Brundage, Murati, Mayer, Welinder, McGrew,
  Amodei, McCandlish, Sutskever, and Zaremba]{copilot}
Chen, M., Tworek, J., Jun, H., Yuan, Q., Pinto, H. P. d.~O., Kaplan, J.,
  Edwards, H., Burda, Y., Joseph, N., Brockman, G., Ray, A., Puri, R., Krueger,
  G., Petrov, M., Khlaaf, H., Sastry, G., Mishkin, P., Chan, B., Gray, S.,
  Ryder, N., Pavlov, M., Power, A., Kaiser, L., Bavarian, M., Winter, C.,
  Tillet, P., Such, F.~P., Cummings, D., Plappert, M., Chantzis, F., Barnes,
  E., Herbert-Voss, A., Guss, W.~H., Nichol, A., Paino, A., Tezak, N., Tang,
  J., Babuschkin, I., Balaji, S., Jain, S., Saunders, W., Hesse, C., Carr,
  A.~N., Leike, J., Achiam, J., Misra, V., Morikawa, E., Radford, A., Knight,
  M., Brundage, M., Murati, M., Mayer, K., Welinder, P., McGrew, B., Amodei,
  D., McCandlish, S., Sutskever, I., and Zaremba, W.
\newblock Evaluating large language models trained on code, 2021.
\newblock URL \url{https://arxiv.org/abs/2107.03374}.

\bibitem[Chowdhery et~al.(2022)Chowdhery, Narang, Devlin, Bosma, Mishra,
  Roberts, Barham, Chung, Sutton, Gehrmann, et~al.]{chowdhery2022palm}
Chowdhery, A., Narang, S., Devlin, J., Bosma, M., Mishra, G., Roberts, A.,
  Barham, P., Chung, H.~W., Sutton, C., Gehrmann, S., et~al.
\newblock {PaLM}: Scaling language modeling with pathways.
\newblock \emph{arXiv preprint arXiv:2204.02311}, 2022.

\bibitem[Creswell et~al.(2022)Creswell, Shanahan, and
  Higgins]{creswell2022selection}
Creswell, A., Shanahan, M., and Higgins, I.
\newblock Selection-inference: Exploiting large language models for
  interpretable logical reasoning.
\newblock \emph{arXiv preprint arXiv:2205.09712}, 2022.

\bibitem[Gao et~al.(2022)Gao, Madaan, Zhou, Alon, Liu, Yang, Callan, and
  Neubig]{gao2022pal}
Gao, L., Madaan, A., Zhou, S., Alon, U., Liu, P., Yang, Y., Callan, J., and
  Neubig, G.
\newblock Pal: Program-aided language models.
\newblock \emph{arXiv preprint arXiv:2211.10435}, 2022.

\bibitem[Goyal \& Bengio(2020)Goyal and Bengio]{bengio2}
Goyal, A. and Bengio, Y.
\newblock Inductive biases for deep learning of human cognition.
\newblock \emph{arXiv preprint arXiv:2011.15091}, 2020.

\bibitem[Kahneman(2011)]{kahneman}
Kahneman, D.
\newblock \emph{Thinking, fast and slow}.
\newblock Macmillan, 2011.

\bibitem[Khot et~al.(2022)Khot, Trivedi, Finlayson, Fu, Richardson, Clark, and
  Sabharwal]{decomp}
Khot, T., Trivedi, H., Finlayson, M., Fu, Y., Richardson, K., Clark, P., and
  Sabharwal, A.
\newblock Decomposed prompting: A modular approach for solving complex tasks.
\newblock \emph{arXiv preprint arXiv:2210.02406}, 2022.

\bibitem[Kojima et~al.(2022)Kojima, Gu, Reid, Matsuo, and
  Iwasawa]{kojima2022large}
Kojima, T., Gu, S.~S., Reid, M., Matsuo, Y., and Iwasawa, Y.
\newblock Large language models are zero-shot reasoners.
\newblock \emph{arXiv preprint arXiv:2205.11916}, 2022.

\bibitem[Li et~al.(2022{\natexlab{a}})Li, Choi, Chung, Kushman, Schrittwieser,
  Leblond, Eccles, Keeling, Gimeno, Dal~Lago, et~al.]{alphacode}
Li, Y., Choi, D., Chung, J., Kushman, N., Schrittwieser, J., Leblond, R.,
  Eccles, T., Keeling, J., Gimeno, F., Dal~Lago, A., et~al.
\newblock Competition-level code generation with alphacode.
\newblock \emph{Science}, 378\penalty0 (6624):\penalty0 1092--1097,
  2022{\natexlab{a}}.

\bibitem[Li et~al.(2022{\natexlab{b}})Li, Lin, Zhang, Fu, Chen, Lou, and
  Chen]{li2022advance}
Li, Y., Lin, Z., Zhang, S., Fu, Q., Chen, B., Lou, J.-G., and Chen, W.
\newblock On the advance of making language models better reasoners.
\newblock \emph{arXiv preprint arXiv:2206.02336}, 2022{\natexlab{b}}.

\bibitem[Liu et~al.(2021)Liu, Shen, Zhang, Dolan, Carin, and
  Chen]{liu2021makes}
Liu, J., Shen, D., Zhang, Y., Dolan, B., Carin, L., and Chen, W.
\newblock What makes good in-context examples for gpt-$3 $?
\newblock \emph{arXiv preprint arXiv:2101.06804}, 2021.

\bibitem[Liu et~al.(2022)Liu, Patwary, Prenger, Prabhumoye, Ping, Shoeybi, and
  Catanzaro]{liu-etal-2022-multi}
Liu, Z., Patwary, M., Prenger, R., Prabhumoye, S., Ping, W., Shoeybi, M., and
  Catanzaro, B.
\newblock Multi-stage prompting for knowledgeable dialogue generation.
\newblock In \emph{Findings of the Association for Computational Linguistics:
  ACL 2022}, pp.\  1317--1337, Dublin, Ireland, May 2022. Association for
  Computational Linguistics.
\newblock \doi{10.18653/v1/2022.findings-acl.104}.
\newblock URL \url{https://aclanthology.org/2022.findings-acl.104}.

\bibitem[Lu et~al.(2022)Lu, Bartolo, Moore, Riedel, and
  Stenetorp]{lu-etal-2022-fantastically}
Lu, Y., Bartolo, M., Moore, A., Riedel, S., and Stenetorp, P.
\newblock Fantastically ordered prompts and where to find them: Overcoming
  few-shot prompt order sensitivity.
\newblock In \emph{Proceedings of the 60th Annual Meeting of the Association
  for Computational Linguistics (Volume 1: Long Papers)}, pp.\  8086--8098,
  Dublin, Ireland, May 2022. Association for Computational Linguistics.
\newblock \doi{10.18653/v1/2022.acl-long.556}.
\newblock URL \url{https://aclanthology.org/2022.acl-long.556}.

\bibitem[Malkin et~al.(2022)Malkin, Wang, and Jojic]{malkin2022coherence}
Malkin, N., Wang, Z., and Jojic, N.
\newblock Coherence boosting: When your pretrained language model is not paying
  enough attention.
\newblock In \emph{Proceedings of the 60th Annual Meeting of the Association
  for Computational Linguistics (Volume 1: Long Papers)}, pp.\  8214--8236,
  2022.

\bibitem[Mialon et~al.(2023)Mialon, Dess{\`\i}, Lomeli, Nalmpantis, Pasunuru,
  Raileanu, Rozi{\`e}re, Schick, Dwivedi-Yu, Celikyilmaz,
  et~al.]{mialon2023augmented}
Mialon, G., Dess{\`\i}, R., Lomeli, M., Nalmpantis, C., Pasunuru, R., Raileanu,
  R., Rozi{\`e}re, B., Schick, T., Dwivedi-Yu, J., Celikyilmaz, A., et~al.
\newblock Augmented language models: a survey.
\newblock \emph{arXiv preprint arXiv:2302.07842}, 2023.

\bibitem[Min et~al.(2022)Min, Lyu, Holtzman, Artetxe, Lewis, Hajishirzi, and
  Zettlemoyer]{min2022rethinking}
Min, S., Lyu, X., Holtzman, A., Artetxe, M., Lewis, M., Hajishirzi, H., and
  Zettlemoyer, L.
\newblock Rethinking the role of demonstrations: What makes in-context learning
  work?
\newblock \emph{arXiv preprint arXiv:2202.12837}, 2022.

\bibitem[Nye et~al.(2021)Nye, Andreassen, Gur-Ari, Michalewski, Austin, Bieber,
  Dohan, Lewkowycz, Bosma, Luan, et~al.]{nye2021show}
Nye, M., Andreassen, A.~J., Gur-Ari, G., Michalewski, H., Austin, J., Bieber,
  D., Dohan, D., Lewkowycz, A., Bosma, M., Luan, D., et~al.
\newblock Show your work: Scratchpads for intermediate computation with
  language models.
\newblock \emph{arXiv preprint arXiv:2112.00114}, 2021.

\bibitem[OpenAI(2023)]{openai2023gpt4}
OpenAI.
\newblock Gpt-4 technical report, 2023.

\bibitem[Ozturkler et~al.(2022)Ozturkler, Malkin, Wang, and Jojic]{thinksum}
Ozturkler, B., Malkin, N., Wang, Z., and Jojic, N.
\newblock Thinksum: Probabilistic reasoning over sets using large language
  models, 2022.
\newblock URL \url{https://arxiv.org/abs/2210.01293}.

\bibitem[Paranjape et~al.(2023)Paranjape, Lundberg, Singh, Hajishirzi,
  Zettlemoyer, and Ribeiro]{paranjape2023art}
Paranjape, B., Lundberg, S., Singh, S., Hajishirzi, H., Zettlemoyer, L., and
  Ribeiro, M.~T.
\newblock Art: Automatic multi-step reasoning and tool-use for large language
  models.
\newblock \emph{arXiv preprint arXiv:2303.09014}, 2023.

\bibitem[Parisi et~al.(2022)Parisi, Zhao, and Fiedel]{parisi2022talm}
Parisi, A., Zhao, Y., and Fiedel, N.
\newblock Talm: Tool augmented language models.
\newblock \emph{arXiv preprint arXiv:2205.12255}, 2022.

\bibitem[Peng et~al.(2023)Peng, Kalliamvakou, Cihon, and
  Demirer]{copilot_study}
Peng, S., Kalliamvakou, E., Cihon, P., and Demirer, M.
\newblock The impact of ai on developer productivity: Evidence from github
  copilot, 2023.
\newblock URL \url{https://arxiv.org/abs/2302.06590}.

\bibitem[Press et~al.(2022)Press, Zhang, Min, Schmidt, Smith, and
  Lewis]{press2022measuring}
Press, O., Zhang, M., Min, S., Schmidt, L., Smith, N.~A., and Lewis, M.
\newblock Measuring and narrowing the compositionality gap in language models.
\newblock \emph{arXiv preprint arXiv:2210.03350}, 2022.

\bibitem[Rae et~al.(2021)Rae, Borgeaud, Cai, Millican, Hoffmann, Song,
  Aslanides, Henderson, Ring, Young, et~al.]{rae2021scaling}
Rae, J.~W., Borgeaud, S., Cai, T., Millican, K., Hoffmann, J., Song, F.,
  Aslanides, J., Henderson, S., Ring, R., Young, S., et~al.
\newblock Scaling language models: Methods, analysis \& insights from training
  {Gopher}.
\newblock \emph{arXiv preprint arXiv:2112.11446}, 2021.

\bibitem[Schick et~al.(2023)Schick, Dwivedi-Yu, Dess{\`\i}, Raileanu, Lomeli,
  Zettlemoyer, Cancedda, and Scialom]{schick2023toolformer}
Schick, T., Dwivedi-Yu, J., Dess{\`\i}, R., Raileanu, R., Lomeli, M.,
  Zettlemoyer, L., Cancedda, N., and Scialom, T.
\newblock Toolformer: Language models can teach themselves to use tools.
\newblock \emph{arXiv preprint arXiv:2302.04761}, 2023.

\bibitem[Schuurmans(2023)]{schuurmans2023memory}
Schuurmans, D.
\newblock Memory augmented large language models are computationally universal.
\newblock \emph{arXiv preprint arXiv:2301.04589}, 2023.

\bibitem[Shi et~al.(2023)Shi, Chen, Misra, Scales, Dohan, Chi, Sch{\"a}rli, and
  Zhou]{shi2023large}
Shi, F., Chen, X., Misra, K., Scales, N., Dohan, D., Chi, E., Sch{\"a}rli, N.,
  and Zhou, D.
\newblock Large language models can be easily distracted by irrelevant context.
\newblock \emph{arXiv preprint arXiv:2302.00093}, 2023.

\bibitem[Shwartz et~al.(2020)Shwartz, West, Le~Bras, Bhagavatula, and
  Choi]{shwartz-etal-2020-unsupervised}
Shwartz, V., West, P., Le~Bras, R., Bhagavatula, C., and Choi, Y.
\newblock Unsupervised commonsense question answering with self-talk.
\newblock In \emph{Proceedings of the 2020 Conference on Empirical Methods in
  Natural Language Processing (EMNLP)}, pp.\  4615--4629, Online, November
  2020. Association for Computational Linguistics.
\newblock \doi{10.18653/v1/2020.emnlp-main.373}.
\newblock URL \url{https://aclanthology.org/2020.emnlp-main.373}.

\bibitem[Srivastava et~al.(2022)Srivastava, Rastogi, Rao, Shoeb, Abid, Fisch,
  Brown, Santoro, Gupta, Garriga-Alonso, et~al.]{srivastava2022beyond}
Srivastava, A., Rastogi, A., Rao, A., Shoeb, A. A.~M., Abid, A., Fisch, A.,
  Brown, A.~R., Santoro, A., Gupta, A., Garriga-Alonso, A., et~al.
\newblock Beyond the imitation game: Quantifying and extrapolating the
  capabilities of language models.
\newblock \emph{arXiv preprint arXiv:2206.04615}, 2022.

\bibitem[Suzgun et~al.(2022)Suzgun, Scales, Sch{\"a}rli, Gehrmann, Tay, Chung,
  Chowdhery, Le, Chi, Zhou, et~al.]{suzgun2022challenging}
Suzgun, M., Scales, N., Sch{\"a}rli, N., Gehrmann, S., Tay, Y., Chung, H.~W.,
  Chowdhery, A., Le, Q.~V., Chi, E.~H., Zhou, D., et~al.
\newblock Challenging big-bench tasks and whether chain-of-thought can solve
  them.
\newblock \emph{arXiv preprint arXiv:2210.09261}, 2022.

\bibitem[Turing(1936)]{turing1936a}
Turing, A.~M.
\newblock On computable numbers, with an application to the
  {E}ntscheidungsproblem.
\newblock \emph{Proceedings of the London Mathematical Society}, 2\penalty0
  (42):\penalty0 230--265, 1936.
\newblock URL
  \url{http://www.cs.helsinki.fi/u/gionis/cc05/OnComputableNumbers.pdf}.

\bibitem[Tversky \& Kahneman(1974)Tversky and Kahneman]{tversky1974judgment}
Tversky, A. and Kahneman, D.
\newblock Judgment under uncertainty: Heuristics and biases: Biases in
  judgments reveal some heuristics of thinking under uncertainty.
\newblock \emph{Science}, 185\penalty0 (4157):\penalty0 1124--1131, 1974.

\bibitem[Wang et~al.(2022{\natexlab{a}})Wang, Wei, Schuurmans, Le, Chi, and
  Zhou]{wang2022rationale}
Wang, X., Wei, J., Schuurmans, D., Le, Q., Chi, E., and Zhou, D.
\newblock Rationale-augmented ensembles in language models.
\newblock \emph{arXiv preprint arXiv:2207.00747}, 2022{\natexlab{a}}.

\bibitem[Wang et~al.(2022{\natexlab{b}})Wang, Wei, Schuurmans, Le, Chi, and
  Zhou]{wang2022self}
Wang, X., Wei, J., Schuurmans, D., Le, Q., Chi, E., and Zhou, D.
\newblock Self-consistency improves chain of thought reasoning in language
  models.
\newblock \emph{arXiv preprint arXiv:2203.11171}, 2022{\natexlab{b}}.

\bibitem[Wei et~al.(2022)Wei, Wang, Schuurmans, Bosma, Chi, Le, and
  Zhou]{wei2022chain}
Wei, J., Wang, X., Schuurmans, D., Bosma, M., Chi, E., Le, Q., and Zhou, D.
\newblock Chain of thought prompting elicits reasoning in large language
  models.
\newblock \emph{arXiv preprint arXiv:2201.11903}, 2022.

\bibitem[Yao et~al.(2022)Yao, Zhao, Yu, Du, Shafran, Narasimhan, and
  Cao]{yao2022react}
Yao, S., Zhao, J., Yu, D., Du, N., Shafran, I., Narasimhan, K., and Cao, Y.
\newblock React: Synergizing reasoning and acting in language models.
\newblock \emph{arXiv preprint arXiv:2210.03629}, 2022.

\bibitem[Zelikman et~al.(2022)Zelikman, Wu, and Goodman]{zelikman2022star}
Zelikman, E., Wu, Y., and Goodman, N.~D.
\newblock {STaR}: Bootstrapping reasoning with reasoning.
\newblock \emph{arXiv preprint arXiv:2203.14465}, 2022.

\bibitem[Zhao et~al.(2021)Zhao, Wallace, Feng, Klein, and
  Singh]{zhao2021calibrate}
Zhao, Z., Wallace, E., Feng, S., Klein, D., and Singh, S.
\newblock Calibrate before use: Improving few-shot performance of language
  models.
\newblock In \emph{International Conference on Machine Learning}, pp.\
  12697--12706. PMLR, 2021.

\bibitem[Zhou et~al.(2022)Zhou, Sch{\"a}rli, Hou, Wei, Scales, Wang,
  Schuurmans, Bousquet, Le, and Chi]{zhou2022least}
Zhou, D., Sch{\"a}rli, N., Hou, L., Wei, J., Scales, N., Wang, X., Schuurmans,
  D., Bousquet, O., Le, Q., and Chi, E.
\newblock Least-to-most prompting enables complex reasoning in large language
  models.
\newblock \emph{arXiv preprint arXiv:2205.10625}, 2022.

\end{thebibliography}
\bibliographystyle{icml2021}
\clearpage

\begin{prompt}[t!]
    \caption{Interpreter/Compiler prompt that can execute LCS algorithm. \href{https://platform.openai.com/playground/p/zCvCyEvZv21fQ7944009aFb3?model=code-davinci-002}{Playground link (use 0 temperature)}}
    \label{pr:interpret}
    \begin{scriptsize}
        \begin{verbatim}
Matrix C contains values, e.g:
C[0,0]=1 C[0,1]=6 C[0,2]=11 C[0,3]=16 C[0,4]=21
C[1,0]=2 C[1,1]=7 C[1,2]=12 C[1,3]=17 C[1,4]=22

To query a value:
Show(C[1,2])
<state> C[1,2]=12 </state>

Query an undefined variable
Show(a)
<state> a=0 </state>

To set a value:
C[0,1]:=8
C[0,1] was 6. Now C[0,1]=8.
a:=5
a was 0. Now a=5.

To query multiple variables:
Show(a, C[0:1,0:4])
<state>
a=5
C[0,0]=1 C[0,1]=8 C[0,2]=11 C[0,3]=16 C[0,4]=21
C[1,0]=2 C[1,1]=7 C[1,2]=12 C[1,3]=17 C[1,4]=22
</state>

Program:
N:=1
for i from 0 to N
    C[i,i]=-3   
Execute:
i:=0
C[0,0]:=-3
i:=1
C[1,1]:=-3
Finished with i=N, so done with i loop

Show(C)
<state>
C[0,0]=-3 C[0,1]=8 C[0,2]=11 C[0,3]=16 C[0,4]=21
C[1,0]=2 C[1,1]=-3 C[1,2]=12 C[1,3]=17 C[1,4]=22
</state>

Program:
if a<C[0,4]
    C[0,0]:=5
else
    C[0,1]:=2
Execute:
Check if a<C[0,1]?  a is 5 C[0,1] is 8 Is 5<8?...
  ... Yes. C[0,0]:=5
Done

Show(C[0,0])
<state> C[0,0]=5 </state>

Program:
C[0,2]:=detailed_max(C[0,3],C[0,4])
Execute:
C[0,3] is 16, C[0,4] is 21. C[0,2] is the greater of...
  ...them. C[0,2]:=21
Done

LCS:
Input: TA ATA End of input
LCS Prep:
a[1]=T a[2]=A
b[1]=A b[2]=T b[3]=A
M=2 N=3

LCS program:
Show(a,b,M,N)
for i from 1 to M
    for j from 1 to N
        if a[i]==b[j]
            C[i,j]:=C[i-1,j-1]+1
        else
            C[i,j]:=detailed_max(C[i,j-1],C[i-1,j])
        Show(i, j, M, N, C[0:i,0:N])
Show('END')
Execute:
        \end{verbatim}
    \end{scriptsize}
\end{prompt}
\begin{prompt}[t!]
    \caption{In response to Prompt \ref{pr:interpret}, the model returns the correct LCS execution path shown here}
    \label{pr:respLCS}
    \begin{scriptsize}
        \begin{verbatim}
        
...

Execute:
\end{verbatim}
%\definecolor{shadecolor}{rgb}{.835, .957, .824}
% \definecolor{shadecolor}{rgb}{0.9569, 1.0000, 0.9490}
\definecolor{shadecolor}{rgb}{0.902, 1, 0.902}
\begin{shaded}
% \begin{plainbox}{}{}    
\begin{verbatim}
<state>
a[1]=T a[2]=A
b[1]=A b[2]=T b[3]=A
M=2 N=3
</state>
i:=1
j:=1
Check if a[1]==b[1]?  a[1] is T b[1] is A Is T==A?...
  ... No. C[1,1]:=detailed_max(C[1,0],C[0,1])
  ... C[1,0] is 0, C[0,1] is 0. C[1,1] is the greater of 
  ...them. C[1,1]:=0
<state>
i=1 j=1 M=2 N=3
C[0,0]=0 C[0,1]=0 C[0,2]=0 C[0,3]=0
C[1,0]=0 C[1,1]=0 C[1,2]=0 C[1,3]=0
</state>
j:=2
Check if a[1]==b[2]?  a[1] is T b[2] is T Is T==T?...
  ... Yes. C[1,2]:=C[0,1]+1
  ... C[0,1] is 0. C[1,2]:=1
<state>
i=1 j=2 M=2 N=3
C[0,0]=0 C[0,1]=0 C[0,2]=0 C[0,3]=0
C[1,0]=0 C[1,1]=0 C[1,2]=1 C[1,3]=0
</state>
j:=3
Check if a[1]==b[3]?  a[1] is T b[3] is A Is T==A?...
  ... No. C[1,3]:=detailed_max(C[1,2],C[0,3])
  ... C[1,2] is 1, C[0,3] is 0. C[1,3] is the greater of 
  ...them. C[1,3]:=1
<state>
i=1 j=3 M=2 N=3
C[0,0]=0 C[0,1]=0 C[0,2]=0 C[0,3]=0
C[1,0]=0 C[1,1]=0 C[1,2]=1 C[1,3]=1
</state>
i:=2
j:=1
Check if a[2]==b[1]?  a[2] is A b[1] is A Is A==A?...
  ... Yes. C[2,1]:=C[1,0]+1
  ... C[1,0] is 0. C[2,1]:=1
<state>
i=2 j=1 M=2 N=3
C[0,0]=0 C[0,1]=0 C[0,2]=0 C[0,3]=0
C[1,0]=0 C[1,1]=0 C[1,2]=1 C[1,3]=1
C[2,0]=0 C[2,1]=1 C[2,2]=0 C[2,3]=0
</state>
j:=2
Check if a[2]==b[2]?  a[2] is A b[2] is T Is A==T?...
  ... No. C[2,2]:=detailed_max(C[2,1],C[1,2])
  ... C[2,1] is 1, C[1,2] is 1. C[2,2] is the greater of 
  ...them. C[2,2]:=1
<state>
i=2 j=2 M=2 N=3
C[0,0]=0 C[0,1]=0 C[0,2]=0 C[0,3]=0
C[1,0]=0 C[1,1]=0 C[1,2]=1 C[1,3]=1
C[2,0]=0 C[2,1]=1 C[2,2]=1 C[2,3]=0
</state>
j:=3
Check if a[2]==b[3]?  a[2] is A b[3] is A Is A==A?...
  ... Yes. C[2,3]:=C[1,2]+1
  ... C[1,2] is 1. C[2,3]:=2
<state>
i=2 j=3 M=2 N=3
C[0,0]=0 C[0,1]=0 C[0,2]=0 C[0,3]=0
C[1,0]=0 C[1,1]=0 C[1,2]=1 C[1,3]=1
C[2,0]=0 C[2,1]=1 C[2,2]=1 C[2,3]=2
</state>
<state>
END
</state>
        \end{verbatim}
        \end{shaded}
        % \end{plainbox}
    \end{scriptsize}
\end{prompt}

\begin{prompt*}
    \caption{Full single path execution path prompt using the state description as in the fragmented Prompt \ref{pr:frag}.  \href{https://platform.openai.com/playground/p/NEpblKDY5x5YaRsLLTxXTx4l?model=code-davinci-002}{Playground link (use with 0 temperature)}} 
    \label{pr:BS2}
    \begin{scriptsize}
        \begin{verbatim}
Problem: 2, 3, 1, 5
EXECUTION
    Length of the list: L=4
    Number of pairs: P=3
    a=[2 3 1 5]
    set n_swaps=0. set i=P=3. set swap_flag=true.   
        <state> a=[2 3 1 5] i=3 P=3 n_swaps=0 swap_flag=true </state>
    Since i=3 and P=3, i and P are equal, so this iteration is done, but swap_flag is true, 
    so we need another iteration
    Iteration:
        set swap_flag=false.  set i=0. The state is:
        <state> a=[2 3 1 5] i=0 P=3 n_swaps=0 swap_flag=false </state>
        Since i=0 and P=3, these two are different, so we continue
        a[i]=a[0]=2 a[i+1]=a[1]=3
        Because 2<3 is true we keep state as is and move on by increasing i
        <state> a=[2 3 1 5] i=1 P=3 n_swaps=0 swap_flag=false </state>
        Since i=1 and P=3, these two are different, so we continue
        a[i]=a[1]=3 a[i+1]=a[2]=1
        Because 3<1 is false we set swap_flag=true,increase n_swaps by one, and in a=[2 3 1 5] swap 3 and 1, 
        and increase i, and keep P as is to get
        <state> a=[2 1 3 5] i=2 P=3 n_swaps=1 swap_flag=true </state>
        Since i=2 and P=3, these two are different, so we continue
        a[i]=a[2]=3 a[i+1]=a[3]=5
        Because 3<5 is true we keep state as is and move on by increasing i
        <state> a=[2 1 3 5] i=3 P=3 n_swaps=1 swap_flag=true </state>
        Since i=3 and P=3, these two are equal, so this iteration is done, but swap_flag is true, 
        so we need another iteration
    Iteration:
        set swap_flag=false.  set i=0. The state is:
        <state> a=[2 1 3 5] i=0 P=3 n_swaps=1 swap_flag=false </state>
        Since i=0 and P=3, these two are different, so we continue
        a[i]=a[0]=2 a[i+1]=a[1]=1
        Because 2<1 is false we set swap_flag=true,increase n_swaps by one, and in a=[2 1 3 5] swap 2 and 1, 
        and increase i, and keep P as is to get
        <state> a=[1 2 3 5] i=1 P=3 n_swaps=2 swap_flag=true </state>
        Since i=1 and P=3, these two are different, so we continue
        a[i]=a[1]=2 a[i+1]=a[2]=3
        Because 2<3 is true we keep state as is and move on by increasing i
        <state> a=[1 2 3 5] i=2 P=3 n_swaps=2 swap_flag=true </state>
        Since i=2 and P=3, these two are different, so we continue
        a[i]=a[2]=3 a[i+1]=a[3]=5
        Because 3<5 is true we keep state as is and move on by increasing i
        <state> a=[1 2 3 5] i=3 P=3 n_swaps=2 swap_flag=true </state>
        Since i=3 and P=3, these two are equal, so this iteration is done, but swap_flag is true, 
        so we need another iteration
    Iteration:
        set swap_flag=false.  set i=0. The state is:
        <state> a=[1 2 3 5] i=0 P=3 n_swaps=2 swap_flag=false </state>
        Since i=0 and P=3, these two are different, so we continue
        a[i]=a[0]=1 a[i+1]=a[1]=2
        Because 1<2 is true we keep state as is and move on by increasing i
        <state> a=[1 2 3 5] i=1 P=3 n_swaps=2 swap_flag=false </state>
        Since i=1 and P=3, these two are different, so we continue
        a[i]=a[1]=2 a[i+1]=a[2]=3
        Because 2<3 is true we keep state as is and move on by increasing i
        <state> a=[1 2 3 5] i=2 P=3 n_swaps=2 swap_flag=false </state>
        Since i=2 and P=3, these two are different, so we continue
        a[i]=a[2]=3 a[i+1]=a[3]=5
        Because 3<5 is true we keep state as is and move on by increasing i
        <state> a=[1 2 3 5] i=3 P=3 n_swaps=2 swap_flag=false </state>
        Since i=3 and P=3, these two are equal, so this iteration is done, but swap_flag is false, so we are done
    Final List: 1, 2, 3, 5
    Number of swaps: 2
    END OF EXECUTION
     
Problem: 3, 6, 8, 2, 7
EXECUTION
        \end{verbatim}
    \end{scriptsize}
    \vspace{-0.2cm}
\end{prompt*}

\begin{prompt}[t!]
    \caption{Prompt that triggers execution of the search for the longest substring without repeating characters. \href{https://platform.openai.com/playground/p/0i2eLLI1xnjy9txtx5j4gZgR?model=code-davinci-002}{Playground link (use 0 temperature)}}
    \label{pr:lss}
    \begin{scriptsize}
        \begin{verbatim}
Input: s = c, b, c, a, b, b
START
Unique letters: a, b, c
Define variables last_a=0, last_b=0, last_c=0
Length of sequence s:  L=6
Because L is 6, the needed number of iterations is 6
set st_ind=1
st m_len=0
set i=1
Iteration 1: 
    s(1) is c, so use last_c
    last_c is 0, so  nothing to do here.
    max(m_len,  i-st_ind+1) is max(0, 1-1+1) which is...
    ...max(0,1)=1, so we set m_len=1
    since i is 1, and the letter is c, set last_c=1
    increase i by one
    i=2, st_ind=1, m_len=1, last_a=0, last_b=0, last_c=1
End of iteration 1. But we need to do 6 iterations,...
...so we do another one
Iteration 2:
    s(2) is b, so use last_b
    last_b is 0, so  nothing to do here.
    max(m_len,  i-st_ind+1) is max(1, 2-1+1) which is...
    ...max(1, 2)=2, so we set m_len=2
    since i is 2, and the letter is b, set last_b=2
    increase i by one
    i=3, st_ind=1, m_len=2, last_a=0, last_b=2, last_c=1
End of iteration 2.  But we need to do 6 iterations,...
...so we do another one
Iteration 3:
    s(3) is c, so use last_c
    last_c is greater than 0, so we reason...
    ...max(st_ind, last_c+1) is max(1, 2)=2...
    ...so we set st_ind=2 
    max(m_len,  i-st_ind+1) is max(2, 3-2+1) which is...
    ...max(2, 2)=2, so we set m_len=2
    since i is 3, and the letter s(3) is c, set last_c=3
    increase i by one
    i=4, st_ind=2, m_len=2, last_a=0, last_b=2, last_c=3
End of iteration 2. But we need to do 6 iterations,...
...so we do another one
Iteration 4:
    s(4) is a, so use last_a
    last_a is 0, so  nothing to do here.
    max(m_len,  i-st_ind+1) is max(2, 4-2+1) which is...
    ...max(2, 3)=3, so we set m_len=3
    since i is 4, and the letter s(4) is a, set last_a=4
    increase i by one
    i=5, st_ind=2, m_len=3, last_a=4, last_b=2, last_c=3
End of iteration 4. But we need to do 6 iterations,...
...so we do another one
Iteration 5:
    s(5) is b, so use last_b
    last_b is greater than 0, so we reason...
    ...max(st_ind, last_b+1) is max(2, 2+1) which is...
    ...max(2, 3)=3 so we set st_ind=3 
    max(m_len,  i-st_ind+1) is max(3, 5-3+1) which is...
    ...max(3, 3)=3, so we set m_len=3
    since i is 5, and the letter s(5) is b, set last_b=5
    increase i by one
    i=6, st_ind=3, m_len=3, last_a=4, last_b=5, last_c=3
End of iteration 5. But we need to do 6 iterations,...
...so we do another one
Iteration 6:
    s(6) is b, so use last_b
    last_b is greater than 0, so we reason...
    ...max(st_ind, last_b+1) is max(3, 5+1) which is...
    ...max(3, 6)=6 so we set st_ind=6 
    max(m_len,  i-st_ind+1) is max(3, 6-6+1) which is...
    ...max(3, 1)=3, so we set m_len=3
    since i is 6, and the letter s(6) is b, set last_b=6
    increase i by one
    i=7, st_ind=6, m_len=3, last_a=4, last_b=6, last_c=3
End of iteration 6. We needed to do 6 iterations,...
...so we are done

The solution is: m_len=3
END

Input: s = p, w, w, k, e, p, z
START
        \end{verbatim}
    \end{scriptsize}
\end{prompt}

\begin{prompt}[t!]
    \caption{Prompt that triggers evaluation of parentheses using a stack. \href{https://platform.openai.com/playground/p/xovdKeaBx0Gq7hDmSfoUsYEg?model=code-davinci-002}{Full prompt in playground}, and \href{https://platform.openai.com/playground/p/yUfZNSxZCn6nqryido2Tdeus?model=code-davinci-002}{Prompt as here (with 10 steps missing) in playground}. Meant to be used with 0 temperature.}
    \label{pr:para}
    \begin{scriptsize}
        \begin{verbatim}
input: ) [ { } ] ( { } ) [ ( { } ) ] } {
input wriritten as a sequence of symbols:
s= ')', '[', '{', '}', ']', '(', '{','}', ')', '[',
'(', '{', '}', ')',']', '}', '{'
length(s)= 17
stack is initialized as empty
i=0
there is nothing in stack, so push s(0)= ')' on stack
stack= )
are the last two symbols an open and a closed
parenthesis of the same type? No. Stack stays same. 
i=1
we push s(1)='[' on the stack
stack= ) [
are the last two symbols an open and a closed
parenthesis of the same type? No. Stack stays same. 
i=2
we push s(2)='{' to the stack
stack= ) [ {
are the last two symbols an open and a closed
parenthesis of the same type? No. Stack stays same. 
i=3
we push s(3)='}' to the stack
stack= ) [ { }
are the last two symbols an open and a closed
parenthesis of the same type? Yes, they are { }, 
opening then closing. 
We pop the last two symbols from the stack.
stack= ) [
i=4
we push s(4)=']' to the stack
stack= ) [ ]
are the last two symbols an open and a closed
parenthesis of the same type? Yes, they are [ ], 
opening then closing. 
We pop the last two symbols from the stack
stack= )
i=5
we push s(5)='(' to the stack
stack= ) (

...

i=15
we push s(15)='}' to the stack
stack= ) }
are the last two symbols an open and a closed
parenthesis of the same type? No. Stack stays same.
i=16
we push s(16)='{' to the stack
stack= ) } {
are the last two symbols an open and a closed
parenthesis of the same type? No. Stack stays same.
i=17
we have reached the end of the input string. 
If the stack has some parenthesis left in it, 
the sequence is invalid, otherwise, 
if the stack is empty, it is valid.
Sequence is: invalid
END

input:
        \end{verbatim}
    \end{scriptsize}
\end{prompt}

\begin{prompt*}
    \caption{A prompt (white background) for translating logical deduction puzzles from BIG-bench into a Wolfram language (Mathematica) call. The response from GPT-3 Codex is shown green-shaded. \href{https://platform.openai.com/playground/p/n1Uo1tpHK7plGTAqYpZyydvu?model=code-davinci-002}{Playground link (use 0 temperature)}}
    \label{pr:wolfram}
    \begin{scriptsize}
        \begin{verbatim}
Puzzle: The following objects need to be ordered. obj1 is the biggest. obj2 is smaller than obj3. 
obj1 is bigger than obj2.

Question: Which object is the biggest?

Parsing step:
    Items: obj1, obj2, obj3
    Numbe of items: 3
    Statement 1: obj1 is the biggest.
    Statement 2: obj2 is smaller than obj3.
    Statement 3: obj1 is bigger than obj2.

Scoring identification step:
     Scores will refer to size. 
     Since we have 3 items, let's assume that the biggest gets a score  of 3 pounds and the smallest gets
     the score of 1 pound.

Translation step:
    Available variable names: x, y, z, a, b, c
    Map item scores of 'obj1', 'obj2', 'obj3' to variable names x, y, z
    obj1 score is x; obj2 score is y; obj3 is z;
    Statement 1: 'x' is the biggest.
    Statement 2: 'y' is smaller than 'z'.
    Statement 3:  'x' is bigger than 'y'.
    Domain constraints: 1<=x<=3, 1<=y<=3, 1<=z<=3
    Statemets as a set of inequalities or equalities: x==3, y<z, x>y
    All varaibles are different: x!=y, x!=z, y!=z
    

Call to WolframAlpha:
    Solve[{1<=x<=3, 1<=y<=3, 1<=z<=3, x==3, y<z, x>y, x!=y, x!=z, y!=z }, {x, y, z}, Integers]
    
END

Puzzle: A fruit stand sells five fruits: kiwis, pears, peaches, loquats, and apples. 
The loquats are more expensive than the kiwis. The pears are more expensive than the peaches. 
The apples are more expensive than the loquats. The apples are less expensive than the peaches.

Question: Which is the most expensive?
\end{verbatim}
%\definecolor{shadecolor}{rgb}{.835, .957, .824}
% \definecolor{shadecolor}{rgb}{0.9569, 1.0000, 0.9490}
\definecolor{shadecolor}{rgb}{0.902, 1, 0.902}
\begin{shaded}
\begin{verbatim}

Parsing step:
    Items: kiwis, pears, peaches, loquats, apples
    Numbe of items: 5
    Statement 1: The loquats are more expensive than the kiwis.
    Statement 2: The pears are more expensive than the peaches.
    Statement 3: The apples are more expensive than the loquats.
    Statement 4: The apples are less expensive than the peaches.

Scoring identification step:
     Scores will refer to price. 
     Since we have 5 items, let's assume that the most expensive gets a score  of 5 dollars and the least expensive
     gets the score of 1 dollar.

Translation step:
    Available variable names: x, y, z, a, b, c
    Map item scores of 'kiwis', 'pears', 'peaches', 'loquats', 'apples' to variable names x, y, z, a, b
    kiwis score is x; pears score is y; peaches is z; loquats is a; apples is b;
    Statement 1: 'a' is more expensive than 'x'.
    Statement 2: 'y' is more expensive than 'z'.
    Statement 3:  'b' is more expensive than 'a'.
    Statement 4:  'b' is less expensive than 'z'.
    Domain constraints: 1<=x<=5, 1<=y<=5, 1<=z<=5, 1<=a<=5, 1<=b<=5
    Statemets as a set of inequalities or equalities: a>x, y>z, b>a, b<z
    All varaibles are different: x!=y, x!=z, x!=a, x!=b, y!=z, y!=a, y!=b, z!=a, z!=b, a!=b
    

Call to WolframAlpha:
    Solve[{1<=x<=5, 1<=y<=5, 1<=z<=5, 1<=a<=5, 1<=b<=5, a>x, y>z, b>a, b<z, x!=y, x!=z, x!=a, x!=b, y!=z, y!=a, 
    y!=b, z!=a, z!=b, a!=b}, {x, y, z, a, b}, Integers]
\end{verbatim}
\end{shaded}

    \end{scriptsize}
\end{prompt*}

\begin{prompt*}
    \caption{A zero-shot code prompt (white background) which, when temperature is carefully chosen may trigger code execution (shaded green) by Codex. Although the sequence gets sorted, the execution goes awry after that. Importantly, the number of swaps is not tracked, even though the code contains appropriate instructions. In contrast, IRSA prompts consistently and repeatedly trigger correct program execution. See also Prompts \ref{pr:GPT4steps1} and \ref{pr:GPT4steps2} for examples of inconsistent behaviour of GPT-4 with zero-shot code prompting.}
    \label{pr:zeroBBScode}
    \begin{scriptsize}
        \begin{verbatim}
# Here is the code for Bubble Sort algorithm that keeps track of the number of needed swaps:
def bubbleSort(arr): 
    n = len(arr) 
    # initiate the number of swaps to 0
    n_swaps=0; 
    # Traverse through all array elements 
    for i in range(n): 
  
        # Last i elements are already in place 
        for j in range(0, n-i-1): 
  
            # traverse the array from 0 to n-i-1 
            # Swap if the element found is greater 
            # than the next element 
            if arr[j] > arr[j+1] : 
                arr[j], arr[j+1] = arr[j+1], arr[j] 
                n_swaps = n_swaps+1

Problem: 8, 7, 6, 4, 9
EXECUTION:
\end{verbatim}
%\definecolor{shadecolor}{rgb}{.835, .957, .824}
% \definecolor{shadecolor}{rgb}{0.9569, 1.0000, 0.9490}
\definecolor{shadecolor}{rgb}{0.902, 1, 0.902}
\begin{shaded}
\begin{verbatim}
    i=0
    j=0
    8 > 7
    7, 8, 6, 4, 9
    j=1
    8 > 6
    7, 6, 8, 4, 9
    j=2
    8 > 4
    7, 6, 4, 8, 9
    j=3
    8 > 9
    7, 6, 4, 8, 9
    j=4
    i=1
    j=0
    7 > 6
    6, 7, 4, 8, 9
    j=1
    7 > 4
    6, 4, 7, 8, 9
    j=2
    7 > 8
    6, 4, 7, 8, 9
    j=3
    i=2
    j=0
    6 > 4
    4, 6, 7, 8, 9
    j=1
    6 > 7
    4, 6, 7, 8, 9
    j=2
    i=3
    j=0
    4 > 6
    4, 6, 7, 8, 9
    j=1
    i=4
    j=0
    i=5
    j=0
    i=6
    j=0
    i=7
    j=0
    i=8
    ...
\end{verbatim}
\end{shaded}

    \end{scriptsize}
\end{prompt*}

\begin{prompt*}
    \caption{An example of a baseline prompt design. Prompts with and without the code, shaded in gray, were used. Zero shot prompts omitted the examples, also shaded in gray. Other than the (buggy) algorithm we adopted for the logical deduction puzzle, the algorithms we tested are known to Codex, and we based the codes used in the prompts on the ones generated by Codex itself.}
    \label{pr:FewShotsCode}
    \begin{scriptsize}
    \begin{verbatim}


     # Python3 program to find the length
     # of the longest substring
     # without repeating characters in string s
     # the maximum length of such a substring will be returned in m_len
     \end{verbatim}
     \definecolor{shadecolor}{rgb}{.95, .95, .96}
     %\definecolor{shadecolor}{rgb}{0.975, 0.975, 0.975}
    \begin{shaded}
        \begin{verbatim}
     def longestUniqueSubsttr(s):
         # last index of every character
         last_idx = {}
         m_len = 0
         # starting index of current
         # window to calculate m_len
         start_idx = 0
         for i in range(0, len(s)):   
             # Find the last index of str[i]
             # Update start_idx (starting index of current window)
             # as maximum of current value of start_idx and last
             # index plus 1
             if s[i] in last_idx:
                 start_idx = max(start_idx, last_idx[s[i]] + 1)
      
             # Update result if we get a larger window
             m_len = max(m_len, i-start_idx + 1)
             # Update last index of current char.
             last_idx[s[i]] = i
         return m_len
    \end{verbatim}
    \end{shaded}
    \begin{verbatim}
     What would the algorithm above, which looks for the longest substring with non repeating characters
     compute for a given problem? Use this format:

     Input: s = ...
     START
     The solution is: m_len= ...
     END
     \end{verbatim}
     \definecolor{shadecolor}{rgb}{.95, .95, .96}
    % \definecolor{shadecolor}{rgb}{0.975, 0.975, 0.975}
     \begin{shaded}
         \begin{verbatim}
     Input: s = a, a, c, a, d, b, b
     START
     The solution is: m_len=4
     END
     
     Input: s = b, b, a, a, d, b, b
     START
     The solution is: m_len=3
     END
     
     Input: s = c, a, d, c, b, c, a
     START
     The solution is: m_len=4
     END
     
     Input: s = a, c, b, a, c, d, d
     START
     The solution is: m_len=4
     END
     
     Input: s = b, b, d, d, a, d, b
     START
     The solution is: m_len=3
     END
\end{verbatim}
\end{shaded}
\begin{verbatim}
     Input: s = a, c, d, c, a, c, a
     START   
\end{verbatim}
    \end{scriptsize}
\end{prompt*}

\begin{prompt*}
    \caption{Given the non-shaded text as a prompt, GPT-4 (0 temperature) returns the green shaded response.}
    \label{pr:GPT4savant}
    \begin{scriptsize}
        \begin{verbatim}
     We need to compute the longest common subsequence for two sequences
     
     s1=bccba
     s2=ccaa
     
     using the dynamic programming algorithm. Show the python code for the algorithm, and then execute it. Finally,
     output the length of the longest common subsequence bracketed with <answer> and </answer>.
\end{verbatim}
%\definecolor{shadecolor}{rgb}{.835, .957, .824}
% \definecolor{shadecolor}{rgb}{0.9569, 1.0000, 0.9490}
\definecolor{shadecolor}{rgb}{0.902, 1, 0.902}
\begin{shaded}
\begin{verbatim}
    Here's the Python code for the dynamic programming algorithm to find the longest common subsequence:
     
     ```python
     def longest_common_subsequence(s1, s2):
         m = len(s1)
         n = len(s2)
         dp = [[0] * (n + 1) for _ in range(m + 1)]
     
         for i in range(1, m + 1):
             for j in range(1, n + 1):
                 if s1[i - 1] == s2[j - 1]:
                     dp[i][j] = dp[i - 1][j - 1] + 1
                 else:
                     dp[i][j] = max(dp[i - 1][j], dp[i][j - 1])
     
         return dp[m][n]
     
     s1 = "bccba"
     s2 = "ccaa"
     lcs_length = longest_common_subsequence(s1, s2)
     print(f"<answer>{lcs_length}</answer>")
     ```   
     
     Now, let us execute the code and find the length of the longest common subsequence:
     
     ```python
     s1 = "bccba"
     s2 = "ccaa"
     lcs_length = longest_common_subsequence(s1, s2)
     print(f"<answer>{lcs_length}</answer>")
     ```  
     
     Output:
     
     ```
     <answer>3</answer>
     ```
\end{verbatim}
\end{shaded}

    \end{scriptsize}
\end{prompt*}

\begin{prompt*}
    \caption{Given the non-shaded text as a prompt, GPT-4 (0 temperature) returns the green shaded response.  The prompt attempts to trigger tracking steps of program execution, but only the initial and final matrix are shown.}
    \label{pr:GPT4steps1}
    \begin{scriptsize}
        \begin{verbatim}
     We need to compute the longest common subsequence for two sequences  
     
     s1=bccba
     s2=ccaa
     
     using the dynamic programming algorithm. Show the python code for the algorithm, and then write down its
     execution with intermediate steps. Finally, output the length of the longest common subsequence bracketed
     with <answer> and </answer>.

\end{verbatim}
%\definecolor{shadecolor}{rgb}{.835, .957, .824}
% \definecolor{shadecolor}{rgb}{0.9569, 1.0000, 0.9490}
\definecolor{shadecolor}{rgb}{0.902, 1, 0.902}
\begin{shaded}
\begin{verbatim}
    Here's the Python code for the dynamic programming algorithm to find the longest common subsequence:
     
     ```python
     def longest_common_subsequence(s1, s2):
         m = len(s1)
         n = len(s2)
         dp = [[0] * (n + 1) for _ in range(m + 1)]
     
         for i in range(1, m + 1):
             for j in range(1, n + 1):
                 if s1[i - 1] == s2[j - 1]:
                     dp[i][j] = dp[i - 1][j - 1] + 1
                 else:
                     dp[i][j] = max(dp[i - 1][j], dp[i][j - 1])
     
         return dp[m][n]
     
     s1 = "bccba"
     s2 = "ccaa"
     print(longest_common_subsequence(s1, s2))
     ```
     
     Now let us s go through the execution with intermediate steps:
     
     1. Initialize the 'dp' table with dimensions (m+1) x (n+1), where m and n are the lengths of s1 and s2,
     respectively. The table will look like this:
     
        
        0 0 0 0 0
        0 0 0 0 0
        0 0 0 0 0
        0 0 0 0 0
        0 0 0 0 0
        0 0 0 0 0
        
     
     2. Iterate through the table, comparing characters from s1 and s2. Update the table based on the dynamic
     programming formula:
     
        
        0 0 0 0 0
        0 0 1 1 1
        0 1 1 1 1
        0 1 2 2 2
        0 1 2 2 2
        0 1 2 2 3
        
     
     3. The length of the longest common subsequence is in the bottom-right cell of the table, which is 3.
     
     So, the length of the longest common subsequence is <answer>3</answer>.
\end{verbatim}
\end{shaded}

    \end{scriptsize}
\end{prompt*}

\begin{prompt*}
    \caption{Given the same non-shaded text as in Prompt \ref{pr:GPT4steps1} as a prompt, {\bf just for a different input}, GPT-4 (0 temperature) returns the green shaded response that this time does list more (but not all!) intermediate steps. This style of prompting GPT-4 reaches accuracy of $69\%$ on LCS-S, compared to $93\%$ achieved with IRSA in GPT-3.}
    \label{pr:GPT4steps2}
    \begin{scriptsize}
        \begin{verbatim}
     We need to compute the longest common subsequence for two sequences 
     
     s1=aaca
     s2=abab   
     
     using the dynamic programming algorithm. Show the python code for the algorithm, and then write down its
     execution with intermediate steps. Finally, output the length of the longest common subsequence bracketed
     with <answer> and </answer>.
\end{verbatim}
%\definecolor{shadecolor}{rgb}{.835, .957, .824}
% \definecolor{shadecolor}{rgb}{0.9569, 1.0000, 0.9490}
\definecolor{shadecolor}{rgb}{0.902, 1, 0.902}
\begin{shaded}
\begin{verbatim}
    Here is the Python code for the dynamic programming algorithm to compute the longest common subsequence:
     ```python
     def longest_common_subsequence(s1, s2):
         m = len(s1)
         n = len(s2)
         
         dp = [[0]*(n+1) for _ in range(m+1)]
         
         for i in range(m+1):
             for j in range(n+1):
                 if i == 0 or j == 0:
                     dp[i][j] = 0
                 elif s1[i-1] == s2[j-1]:
                     dp[i][j] = dp[i-1][j-1] + 1
                 else:
                     dp[i][j] = max(dp[i-1][j], dp[i][j-1])
         
         return dp[m][n]
     
     s1 = "aaca"
     s2 = "abab"
     result = longest_common_subsequence(s1, s2)
     print(result)
     ```
     Here is the execution with intermediate steps:
     
     1. Initialize the 'dp' table with dimensions (m+1) x (n+1). For the given sequences, it will look like:
        
        0 0 0 0 0
     0  0 0 0 0 0
     0  0 0 0 0 0
     0  0 0 0 0 0
     0  0 0 0 0 0    
     
     2. Fill in the 'dp' table by iterating through each cell and apply the algorithm rules:
          
     For i, j = 1, 1: s1[i-1] != s2[j-1], so dp[i][j] = max(dp[i-1][j], dp[i][j-1]) = max(0, 0) = 0
        0 0 0 0 0
     0  0 0 0 0 0
     0  0 0 0 0 0
     0  0 0 0 0 0
     0  0 0 0 0 0
     
     For i, j = 1, 2: s1[i-1] != s2[j-1], so dp[i][j] = max(dp[i-1][j], dp[i][j-1]) = max(0, 0) = 0
        0 0 0 0 0
     0  0 0 0 0 0
     0  0 0 0 0 0
     0  0 0 0 0 0
     0  0 0 0 0 0
     
     For i, j = 1, 3: s1[i-1] == s2[j-1], so dp[i][j] = dp[i-1][j-1] + 1 = 0 + 1 = 1  
        0 0 0 0 0
     0  0 0 0 0 0
     0  0 0 1 0 0
     0  0 0 0 0 0
     0  0 0 0 0 0  
     
     Continue filling in the table with the same rules:
        
        0 0 0 0 0
     0  0 0 0 0 0
     0  0 0 1 0 0
     0  0 1 1 1 0
     0  0 1 1 2 0
          
     3. The value in the bottom-right cell contains the length of the longest common subsequence:
        
     <answer>2</answer>
\end{verbatim}
\end{shaded}

    \end{scriptsize}
\end{prompt*}
\end{document}